\RequirePackage{tikz}

\documentclass[default,iicol]{sn-jnl}

\theoremstyle{thmstyleone}%

\theoremstyle{thmstyletwo}%
\theoremstyle{thmstylethree}%
\usepackage{fancyhdr}
\usepackage{adjustbox}
\usepackage{amsmath}
\usepackage{amsfonts}
\usepackage{amssymb}
\usepackage{array}
\usepackage{booktabs}
\usepackage{comment}
\usepackage{cleveref}

\usepackage{microtype}
\usepackage{multirow}
\usepackage{multicol}
\usepackage{threeparttable}
\usepackage{makecell}

\usepackage[utf8]{inputenc}
\usepackage{program}
\usepackage{color}
\usepackage{pgfplots} 
\pgfplotsset{compat=1.17}
\usepackage{graphicx}
\usepackage[figuresright]{rotating}

\usepackage{hyperref}
\usepackage[hyphenbreaks]{breakurl}
\usepackage{float}
\usepackage{subfigure}
\usepackage{pgf-pie} 

\newcommand{\ie}{\textit{i}.\textit{e}.}
\newcommand{\eg}{\textit{e}.\textit{g}.}

\begin{document}

\title[Article Title]{Deep Learning for Video-Text Retrieval: a Review}

\author[1]{\fnm{Cunjuan} \sur{Zhu}}\email{Zhucunjuan@163.com}

\author[1]{\fnm{Qi} \sur{Jia}}\email{jiaqi@dlut.edu.cn}

\author[2]{\fnm{Wei} \sur{Chen}}\email{weichen@nudt.edu.cn}

\author[2]{\fnm{Yanming} \sur{Guo}}\email{guoyanming@nudt.edu.cn}

\author*[1]{\fnm{Yu} \sur{Liu}}\email{liuyu8824@dlut.edu.cn}

\affil[1]{Dalian University of Technology, China}

\affil[2]{National University of Defense Technology, China}


\abstract{Video-Text Retrieval (VTR) aims to search for the most relevant video related to the semantics in a given sentence, and vice versa. In general, this retrieval task is composed of four successive steps: video and textual feature representation extraction, feature embedding and matching, and objective functions. In the last, a list of samples retrieved from the dataset is ranked based on their matching similarities to the query. In recent years, significant and flourishing progress has been achieved by deep learning techniques, however, VTR is still a challenging task due to the problems like how to learn an efficient spatial-temporal video feature and how to narrow the cross-modal gap. In this survey, we review and summarize over 100 research papers related to VTR, demonstrate state-of-the-art performance on several commonly benchmarked datasets, and discuss potential challenges and directions, with the expectation to provide some insights for researchers in the field of video-text retrieval.
}


\maketitle

\thispagestyle{fancy}
\fancyhead{}
\rhead{}
\lfoot{}

\cfoot{\quad}
\renewcommand{\headrulewidth}{0pt}
\renewcommand{\footrulewidth}{0pt}
\pagestyle{empty}

\section{Introduction}\label{sec1}
With the advent of the big data era, video media software (\eg, YouTuBe, TikTok, and Instagram) has become the main way for human beings to learn and entertain. Facing the explosive growth of multimedia information, people urgently need to find efficient ways to quickly search for an item that meets the user's needs. In particular, the number of videos with different lengths has soared rapidly in recent years, whereas it becomes more time-consuming and difficult to find the target video. Motivated by this, one of the most appropriate solutions is to search for the corresponding video by a language sentence, \ie, Video-Text Retrieval (VTR).

\begin{figure}[ht]%
\centering
\includegraphics[width=0.5\textwidth]{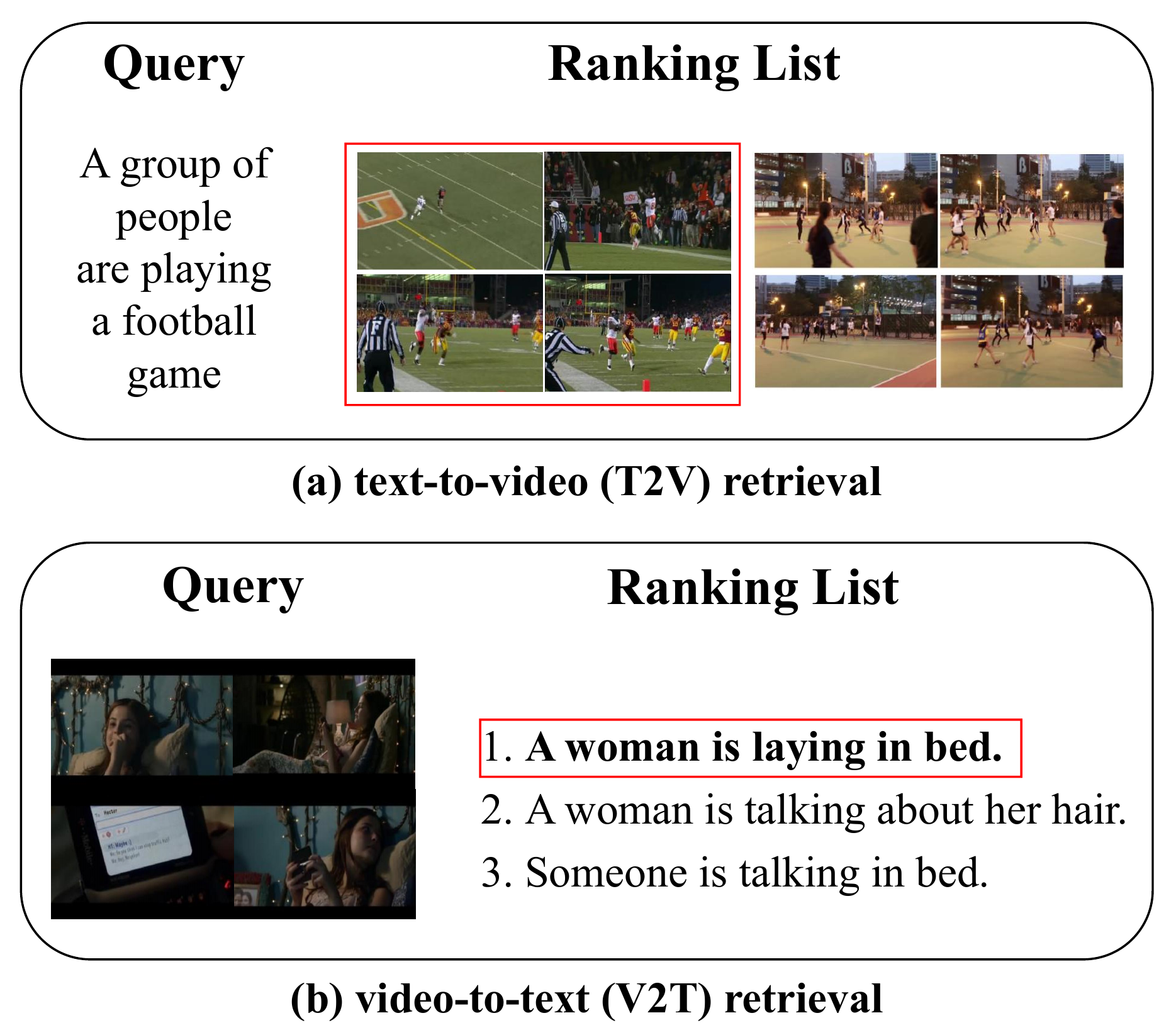}
\caption{Video-text retrieval examples on the MSR-VTT dataset. The red box indicates the item is retrieved correctly.
}\label{Retrieval instance}
\end{figure}

To be more specific, VTR~\cite{mithun2018learning,guo2021ssan,wang2021t2vlad,gao2021clip2tv,luo2021CLIP4clip,fang2021clip2video,ma2022x,zhao2022centerclip} refers to analyze a given sentence so as to find the most matching video from the corpus, and vice versa, as shown in Fig.~\ref{Retrieval instance}. VTR requires content analysis of a large number of video-text pairs, fully excavating multi-modal information involved, and judging whether the two modalities can be aligned. The majority of vision tasks are longstanding and have achieved excellent results in recent decades, including visual classification, object detection, and semantic segmentation. However, VTR is still an emerging task and has not been studied extensively. 

Although the performance of VTR is improving incrementally, there are still some problems and challenges that need further exploration. 
This review aims to revisit deep learning methods for video-text retrieval and provide an extensive summary for researchers who are devoted to this field. In the following of this section, we will describe the generic pipeline for VTR and the major recent advances.

\subsection{Generic Pipeline for VTR}\label{General Framework}
\begin{figure*}[hbt!]
\centering
\includegraphics[width=0.95\textwidth]{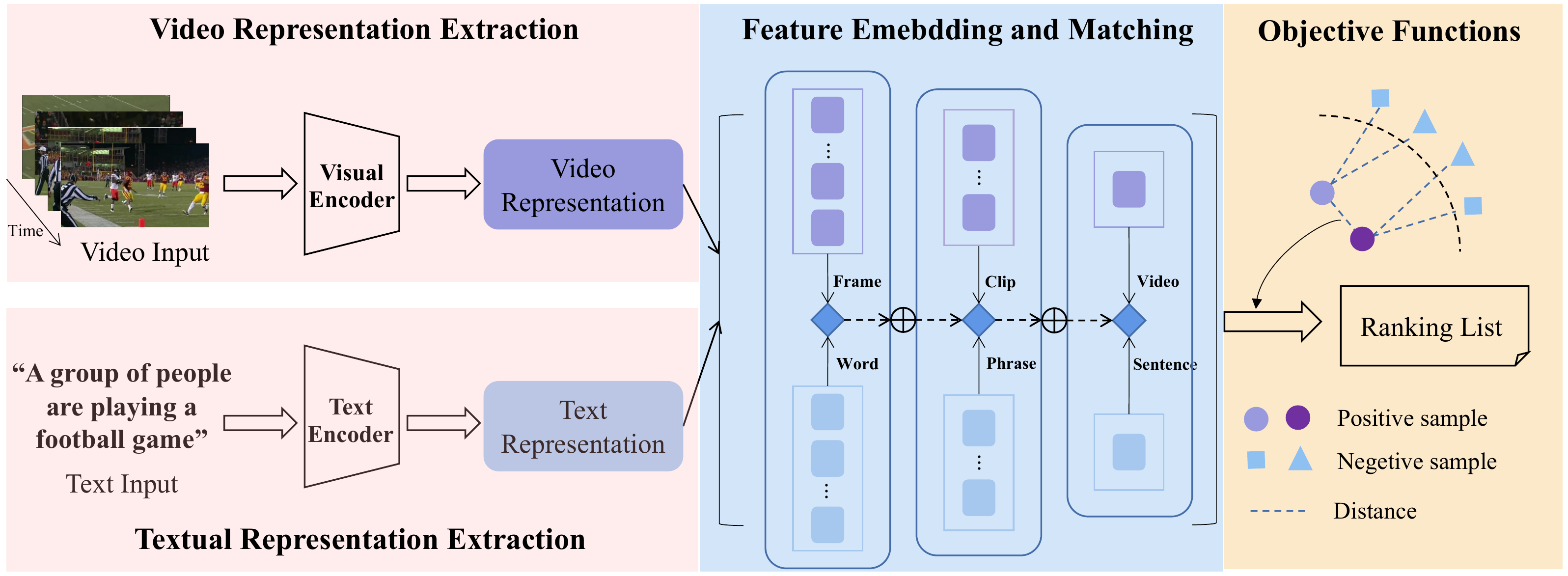}
\caption{Overview of video-text retrieval framework. Given a pair of video and text, the first step is to extract feature representation, followed by feature embedding and matching to compute similarity scores, and the final ranking list is obtained by optimizing the objective functions In part of feature embedding and matching, from left to right, three different granularity matching strategies that vary from fine-grained to coarse-grained can be combined or directly applied separately.
}
\label{framework}
\end{figure*}

In general, the VTR task can be divided into four parts: video representation extraction, textual representation extraction, feature embedding and matching, and objective functions. Figure~\ref{framework} depicts the overview pipeline for video-text retrieval.
\begin{itemize}
    \item [(1)]\textbf{Video representation extraction} is to capture video feature representation. In this survey, we divide these video feature extractors into spatial feature extractors~\cite{hu2018squeeze,guo2021ssan,wu2021hanet} and temporal feature extractors~\cite{qiu2019learning,fang2021clip2video,cheng2021improving,han2021visual} according to the spatio-temporal property. In addition, since multi-modality information is involved in the videos (\eg, motion, audio, or face features), additional experts are typically used to extract features for each modality, and then are aggregated to generate a more comprehensive video representation~\cite{miech2018learning,liu2019use,gabeur2020multi,wang2022hybrid}.
    \item [(2)]\textbf{Textual representation extraction} refers to extract textual features, and the extractors are mainly built upon RNNs~\cite{otani2016learning,wu2021hanet}, CNNs~\cite{miech2018learning,fan2020person}, and Transformer~\cite{ging2020coot,liu2021hit,gabeur2020multi}. Most previous methods~\cite{dong2019dual,chen2020fine} learn and train dual encoders to extract video and textual features separately, but recent work~\cite{radford2021learning} proposes a novel Contrastive Language Image Pre-training (CLIP) model which can learn a multi-modal feature representation in a joint learning fashion. CLIP is composed of an Image Encoder for representing videos and a Language Encoder for textual representations~\cite{luo2021CLIP4clip,gao2021clip2tv,fang2021clip2video}. 
    \item [(3)]\textbf{Feature embedding and matching} aims to map and align extracted video and text representations and compute their similarity scores (also known as compatibility). The core is how to embed the two modalities in a joint feature space and judge whether they match or not. In order to bridge their semantic gap and measure the similarity more accurately, typical approaches can be divided into global, local, and individual matching according to semantic hierarchy. In addition, most works~\cite{wang2021t2vlad,dong2022reading,wang2022hybrid} consider the combination of global and local features. Furthermore,~Min et al.~\cite{min2022hunyuan_tvr} and~Wang et al.~\cite{wang2022disentangled} are dedicated to aligning more fine-grained features, \ie, individual-wise. In addition, other recent studies~\cite{chen2020fine,wu2021hanet,cheng2021improving} divide video and text features into the entity, action, and other levels according to the hierarchical semantic graph.
    \item [(4)]\textbf{Objectives functions} are to train and refine feature representations from the extractors and similarity scores from feature matching. Existing methods define a variety of objective functions, which can be roughly classified as Triplet loss, Contrastive loss, and other loss. Specifically, triplet losses mainly include bi-directional max-margin ranking loss~\cite{gabeur2020multi,wang2021t2vlad} and hinge-based triplet ranking loss~\cite{han2021visual}. InfoNCE~\cite{liu2021hit,gao2021clip2tv,ma2022x} and Dual Softmax Loss~\cite{cheng2021improving} are commonly utilized for contrastive learning \textcolor{black}{objectives}. Additionally, other works introduce knowledge distillation loss~\cite{croitoru2021teachtext}. Moreover, a few studies~\cite{wu2021hanet,fang2021clip2video} choose to combine several loss functions jointly to obtain more optimized results. 
\end{itemize}

\begin{figure*}
\centering
\includegraphics[width=\textwidth]{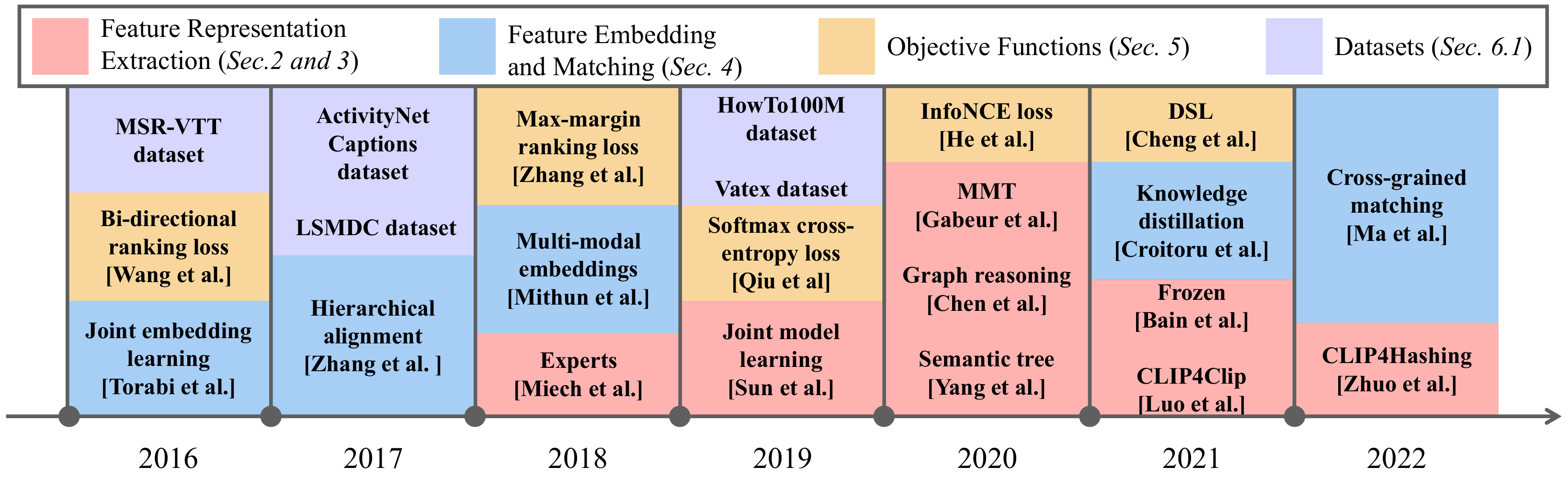}
\caption{Major advances on VTR approaches from 2016 to 2022: the pink boxes show the development of video and text representation extraction~\cite{miech2018learning,sun2019learning,gabeur2020multi,chen2020fine,yang2020tree,bain2021frozen,luo2021CLIP4clip,zhuo2022clip4hashing} described in Sec.~\ref{Video Represetation} and~\ref{Text Representation}; the development of \textcolor{blue}{feature embedding and matching} described in Sec.~\ref{Feature matching} is represented by blue boxes~\cite{torabi2016learning,zhang2018cross,mithun2018learning,croitoru2021teachtext,ma2022x}; orange boxes demonstrate commonly utilized objective functions~\cite{wang2016learning,zhang2018cross,qiu2019learning,he2020momentum,cheng2021improving} in Sec.~\ref{Training Objectives}; the large-scale datasets~\cite{xu2016msr,krishna2017dense,rohrbach2017movie,miech2019howto100m,wang2019vatex} are listed in purple boxes and introduced in 
Sec.~\ref{Datasets}.}\label{Development}
\end{figure*}

\subsection{Major Recent Advances}\label{Major Recent Advances}
Over the past few years, numerous deep learning methods for both text-to-video and video-to-text have been studied continuously. We depict the representative progresses from 2016 to 2022 in Fig.~\ref{Development} from four aspects: feature representation extraction, feature embedding and matching, objectives functions, and datasets.

\begin{itemize}
    \item [(1)]\textbf{Feature Representation Extraction.} In terms of video and text representation extraction, most existing methods utilize CNNs and RNNs to extract features. 
    Since 2018, some studies begin to extract multi-modal representations with different extractors (also known as experts), and then fuse them to obtain the final video features. Besides, different from building a dual encoder to extract video and text features separately, some works propose learning a joint model to capture multi-modal representation since 2019~\cite{sun2019videobert}. In 2020, graph reasoning is introduced to the VTR field for learning textual hierarchical features~\cite{chen2020fine}. Likewise,~Yang et al.~\cite{yang2020tree} apply a latent semantic tree to capture the relations among the words. Since the proposal of BERT~\cite{devlin2018bert} in 2018, Transformer based approaches have achieved a breakthrough for boosting retrieval performance.~Gabeur et al.~\cite{gabeur2020multi} adopt BERT to extract textual features and propose to fuse multi-modal embedding from multiple experts. Since ViT (Vision Transformer) was proposed in 2020, Transformer has become a powerful backbone for recent works like Frozen~\cite{bain2021frozen} and CLIP4Clip~\cite{luo2021CLIP4clip}. Very recently, CLIP4Hashing~\cite{zhuo2022clip4hashing} introduces a hashing encoder that raises a novel idea for feature extraction.

\item [(2)]\textbf{Feature Embedding and Matching.} One of the first works for text-to-video retrieval was in 2016~\cite{torabi2016learning}, which proposed video-language joint embedding learning to map video and text features into a joint space to bridge the cross-modal gap for better feature matching. Linear projection is a commonly used way for feature embedding, and other nonlinear methods such as clustering centers and MLPs are applicable to perform feature matching. Since 2017, Zhang et al.~\cite{zhang2018cross} have begun to consider the impact of different hierarchical semantic information on feature matching, which gradually evolved into global, local and individual levels as the following years progressed. Besides,~Mithun et al.~\cite{mithun2018learning} arise multi-modal embeddings for aligning different modality features. Since the emergence of the cross-encoder model, video and text representations can be concatenated together and fed into a cross-encoder, generating the similarity score via a linear layer. Furthermore, cross-grained similarity calculation~\cite{ma2022x} is first proposed in 2022.

    \item [(3)]\textbf{Objectives functions.} For optimizing feature representations, multiple objective functions are defined for the VTR task, for example, bi-directional ranking loss~\cite{miech2018learning}, max-margin ranking loss~\cite{zhang2018cross}, softmax cross-entropy loss~\cite{qiu2019learning} and InfoNCE loss~\cite{he2020momentum}. Notably, Dual Softmax Loss (DSL)~\cite{cheng2021improving} is proposed in 2021, which achieves promising performance gains in many works~\cite{gao2021clip2tv,luo2021CLIP4clip,min2022hunyuan_tvr}.

    \item [(4)]\textbf{Datasets.} Since 2016, several large-scale benchmarks such as MSR-VTT~\cite{xu2016msr}, ActivityNet Captions~\cite{krishna2017dense}, LSMDC~\cite{rohrbach2017movie} and HowTo100M~\cite{miech2019howto100m} are collected successively to provide the data support for researching the VTR task. Besides, Vatex~\cite{wang2019vatex} proposed a bilingual dataset in 2019.

\end{itemize}

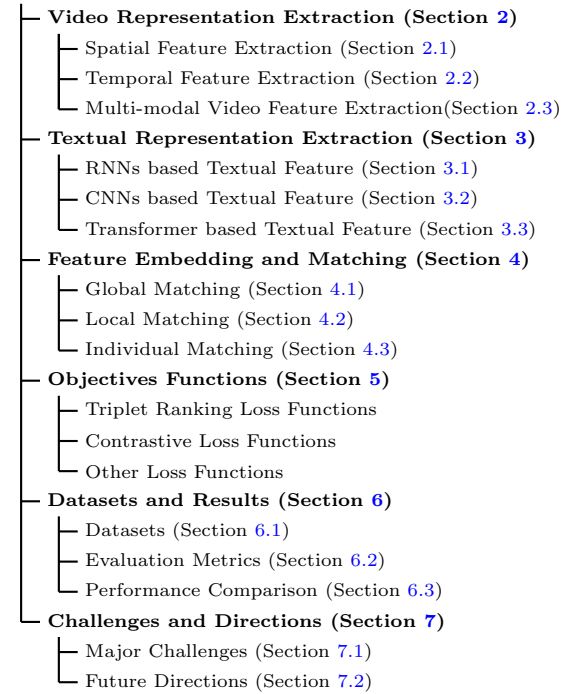
\begin{figure}[!htbp]
\centering
\footnotesize
\begin{tikzpicture}[xscale=0.5, yscale=0.4]

\draw [thick, -] (0, 16.5) -- (0, -4); \node [right] at (-0.5, 17) {\bf Deep Learning for Video-Text Retrieval: a Review};

\draw [thick, -] (0, 16) -- (0.5, 16);\node [right] at (0.5, 16) 
{\bf Video Representation Extraction (Section \ref{Video Represetation})};
\draw [thick, -] (1, 15.5) -- (1, 13);

\draw [thick, -] (1, 15) -- (1.5, 15);\node [right] at (1.5, 15){Spatial Feature Extraction
(Section \ref{Video_spatial})
};
\draw [thick, -] (1, 15.5) -- (1, 13); 

\draw [thick, -] (1, 14) -- (1.5, 14);\node [right] at (1.5, 14) {Temporal Feature Extraction 
(Section \ref{Video_temporal})
};

\draw [thick, -] (1, 13) -- (1.5, 13);\node [right] at (1.5, 13)  {Multi-modal Video Feature Extraction
(Section \ref{video_expert})
};

\draw [thick, -] (0, 12) -- (0.5, 12);\node [right] at (0.5, 12) {\bf Textual Representation Extraction (Section \ref{Text Representation}) };

\draw [thick, -] (1, 11.5) -- (1, 9); 

\draw [thick, -] (1.0, 11) -- (1.5, 11);\node [right] at (1.5, 11) {RNNs based Textual Feature 
(Section \ref{text-RNNs})
};

\draw [thick, -] (1.0, 10) -- (1.5, 10);\node [right] at (1.5, 10) {CNNs based Textual Feature
(Section \ref{text-CNNs})
};

\draw [thick, -] (1.0, 9) -- (1.5, 9);\node [right] at (1.5, 9) {Transformer based Textual Feature
(Section \ref{text-Transformer}) 
};

\draw [thick, -] (0, 8) -- (0.5, 8);\node [right] at (0.5, 8) {\bf Feature Embedding and Matching (Section \ref{Feature matching})};
\draw [thick, -] (1, 7.5) -- (1, 5);

\draw [thick, -] (1, 7) -- (1.5, 7);\node [right] at (1.5, 7) { Global Matching (Section \ref{Feature matching with global}) };

\draw [thick, -] (1, 6) -- (1.5, 6);\node [right] at (1.5, 6) { Local Matching (Section \ref{Feature matching with local}) };
\draw [thick, -] (1, 5) -- (1.5, 5);\node [right] at (1.5, 5) { Individual Matching (Section \ref{Feature matching with individual}) };

\draw [thick, -] (0, 4) -- (0.5, 4);\node [right] at (0.5, 4) {\bf Objectives Functions (Section \ref{Training Objectives})};
\draw [thick, -] (1, 3.5) -- (1, 1);

\draw [thick, -] (1, 3) -- (1.5, 3);\node [right] at (1.5, 3) { Triplet Ranking Loss Functions  };

\draw [thick, -] (1.0, 2) -- (1.5, 2);\node [right] at (1.5, 2) { Contrastive Loss Functions};
\draw [thick, -] (1.0, 1) -- (1.5, 1);\node [right] at (1.5, 1) { Other Loss Functions};

\draw [thick, -] (0, 0) -- (0.5, 0);\node [right] at (0.5, 0) {\bf Datasets and Results (Section \ref{Datasets and Results})};

\draw [thick, -] (1, -0.5) -- (1, -3); 
\draw [thick, -] (1.0, -1) -- (1.5, -1);\node [right] at (1.5, -1) { Datasets (Section \ref{Datasets})};

\draw [thick, -] (1.0, -2) -- (1.5, -2);\node [right] at (1.5, -2) { Evaluation Metrics (Section \ref{Evaluaiton Metrics})};

\draw [thick, -] (1.0, -3) -- (1.5, -3);\node [right] at (1.5, -3) { Performance Comparison (Section \ref{Performance Comparison})};

\draw [thick, -] (0, -4) -- (0.5, -4);\node [right] at (0.5, -4) {\bf Challenges and Directions (Section \ref{Challenges and Directions})};

\draw [thick, -] (1, -4.5) -- (1, -6); 
\draw [thick, -] (1.0, -5) -- (1.5, -5);\node [right] at (1.5, -5) { Major Challenges (Section \ref{key challenges})};

\draw [thick, -] (1.0, -6) -- (1.5, -6);\node [right] at (1.5, -6) { Future Directions (Section \ref{Future Directions and Outlook})};

\end{tikzpicture}
\caption{Structure of the content in this survey.}\label{Pipeline}
\end{figure}

\color{black}

    
\begin{table*}
\centering
\resizebox{\linewidth}{!}{ 
\begin{threeparttable}
\caption{\large Overview of feature extractors for videos and sentences based on deep learning. Each `\checkmark' refers to the network that the corresponding work utilizes. The first column indicates the application of experts, which refers to multi-modal video features. The following two columns extract video features from spatial and temporal characteristics. The fourth column presents the use of textual encoders. The last column lists the published papers and reports their performance results on the MSR-VTT 1k-A split.
}\label{classify}
\begin{tabular}{@{\extracolsep{\fill}}c|cc|cccc|cccc|rlll@{\extracolsep{\fill}}}
\toprule
\multirow{2}{*}{\makecell {Multi-modal video \\ feature (Experts)}} & \multicolumn{2}{@{}c@{}}{Spatial feature extraction} &  \multicolumn{4}{@{}c@{}}{Temporal feature extraction} & \multicolumn{4}{@{}c@{}@{}}{Textual feature extraction} &\multirow{2}{*}{Method-Year-Results} \\\cmidrule(r){2-7} \cmidrule(r){8-11}%
&CNNs & Transf(ViT) &Agg stra. &RNNs &CNNs &Transf &RNNs & CNNs/others & Transf & BERT \\
\midrule

\checkmark &~&~&\checkmark &~ &~ & &~ &\checkmark  &~ &  
&MEE~\cite{miech2018learning}2018-10.1(L)\\
\checkmark &~&~&\checkmark  &~ &~ &~ &~ &\checkmark 
&~ & 
&CE~\cite{liu2019use}2019-20.9\\
\checkmark &~&~&\checkmark  &~ &~ &\checkmark &~ &~ &~ &\checkmark  
&MMT~\cite{gabeur2020multi}2020-24.6\\
\checkmark&~& & &~ &~ &\checkmark &&~ &~ &\checkmark
&HCQ~\cite{wang2022hybrid}2022-25.9\\
\checkmark &~&~&\checkmark  &~ &~ & &~ &~ &~ &\checkmark 
&T2VLAD~\cite{wang2021t2vlad}2021-29.5\\
\checkmark&\checkmark 
&~&~ &~ &\checkmark 
&~ &\checkmark &~ &~ &~  
&Fusion~\cite{mithun2018learning}2018-12.5\\
\checkmark &\checkmark 
&~&~ &\checkmark&\checkmark 
&~ &\checkmark   &\checkmark 
&~ &~  &JsFusion~\cite{yu2018joint}2018-10.2\\
\checkmark &\checkmark 
 &~& &~ &\checkmark 
 &\checkmark  & &~ &\checkmark &
&SUPPORT-SET~\cite{patrick2020support}2020-27.4\\
\checkmark&\checkmark 
&~&  &~ &\checkmark 
&\checkmark &~ &~ &~ &\checkmark  
&HiT~\cite{liu2021hit}2021-28.8\\
\checkmark&\checkmark 
&\checkmark & &~ &\checkmark  
&\checkmark &&~ & &  \checkmark
&VSR-Net~\cite{han2021visual}2021-37.2\\
\checkmark&~&\checkmark 
&~ &~ &\checkmark 
&\checkmark &~ &~ &~ &\checkmark 
&MDMMT~\cite{dzabraev2021mdmmt}2021-38.9\\
\hline
&\checkmark&~&\checkmark &\checkmark &~ & &\checkmark &\checkmark 
&~ &\checkmark  
&Dual Enc.~\cite{dong2019dual}2019-7.7(Mfull)\\
&\checkmark &~&~ &~ &\checkmark & &~ &\checkmark 
&~ &
&HowTo100M~\cite{miech2019howto100m}2019-14.9\\
&\checkmark &~& &\checkmark &~ & &\checkmark   &~ &~ & &TCE~\cite{yang2020tree}2020-17.1\\
&\checkmark &~&\checkmark  &~ &~ & &~ &\checkmark 
&~ & 
&ClipBERT~\cite{lei2021less}2021-22.0\\
&\checkmark& & &\checkmark &~ &~ &\checkmark 
&\checkmark & &  &Ali et al.~\cite{ali2022video}2022-26.0\\
&\checkmark& & &\checkmark  &~ &\checkmark &&~ &~ &\checkmark 
&RIVRL~\cite{dong2022reading}2022-27.9\\
&~&\checkmark & &~ &~ &\checkmark &&~ &
&\checkmark  &Frozen~\cite{bain2021frozen}2021-31.0\\
&~&\checkmark &\checkmark&~ &~ & &&~ & 
&\checkmark &MILES~\cite{ge2022miles}2022-37.7\\
&~&\checkmark &\checkmark  &~ &~ & &&~ & &  \checkmark
&CAMoE~\cite{cheng2021improving}2021-47.3\\
\hline
&~&\checkmark & &~ &\checkmark 
&\checkmark &&~ &\checkmark&  
&MDMMT-2~\cite{kunitsyn2022mdmmt}2022-48.5\\
&~&\checkmark & &~ &~ &\checkmark &&~ &\checkmark & 
&CLIP2Video~\cite{fang2021clip2video}2021-45.6\\
&~&\checkmark & &~ &~ &\checkmark &&~ &\checkmark  &
&CLIP2TV~\cite{gao2021clip2tv}2021-52.9\\
&~&\checkmark & &~ &~ & &&~ &\checkmark &  
&CenterCLIP~\cite{zhao2022centerclip}2022-48.4\\
&~&\checkmark &\checkmark  &\checkmark & &\checkmark & &~ &\checkmark   & 
&CLIP4Clip~\cite{luo2021CLIP4clip}2021-44.5\\
&~&\checkmark &\checkmark  &~ &~ &\checkmark &&~ &\checkmark&  
&HunYuan\-tvr~\cite{min2022hunyuan_tvr}2022-55.0\\
&~&\checkmark & &~ &~ & &&~ &\checkmark &  
&DRL~\cite{wang2022disentangled}2022-53.3\\
&~&\checkmark & &~ &~ &\checkmark &&~ &\checkmark&  
&X-CLIP~\cite{ma2022x}2022-49.3\\
&~&\checkmark &\checkmark  &~ &~ & &&~ &\checkmark &
&X-Pool~\cite{gorti2022x}2022-46.9\\
\botrule
\end{tabular}
\end{threeparttable}}
\end{table*}
    
The rest of this survey paper is organized as shown in Fig.~\ref{Pipeline}.
In Section~\ref{Video Represetation}, we introduce how to extract video features.
Section~\ref{Text Representation} gives an introduction to existing methods for textual feature extraction. In Section ~\ref{Feature matching}, we present how to embed video and text representation into a common space for feature matching. Section~\ref{Training Objectives} describes several objective loss functions widely utilized to train feature representation and matching. In Section~\ref{Datasets and Results}, we summarize existing large-scale VTR datasets and compare the results of recent methods on these datasets. Afterward, Section~\ref{Challenges and Directions} introduces the key challenges that are still 
remaining and discusses several potential directions in near future. Finally, in Section~\ref{conclusion} we draw our conclusions.

\begin{figure*}
\centering
\includegraphics[width=0.8\textwidth]{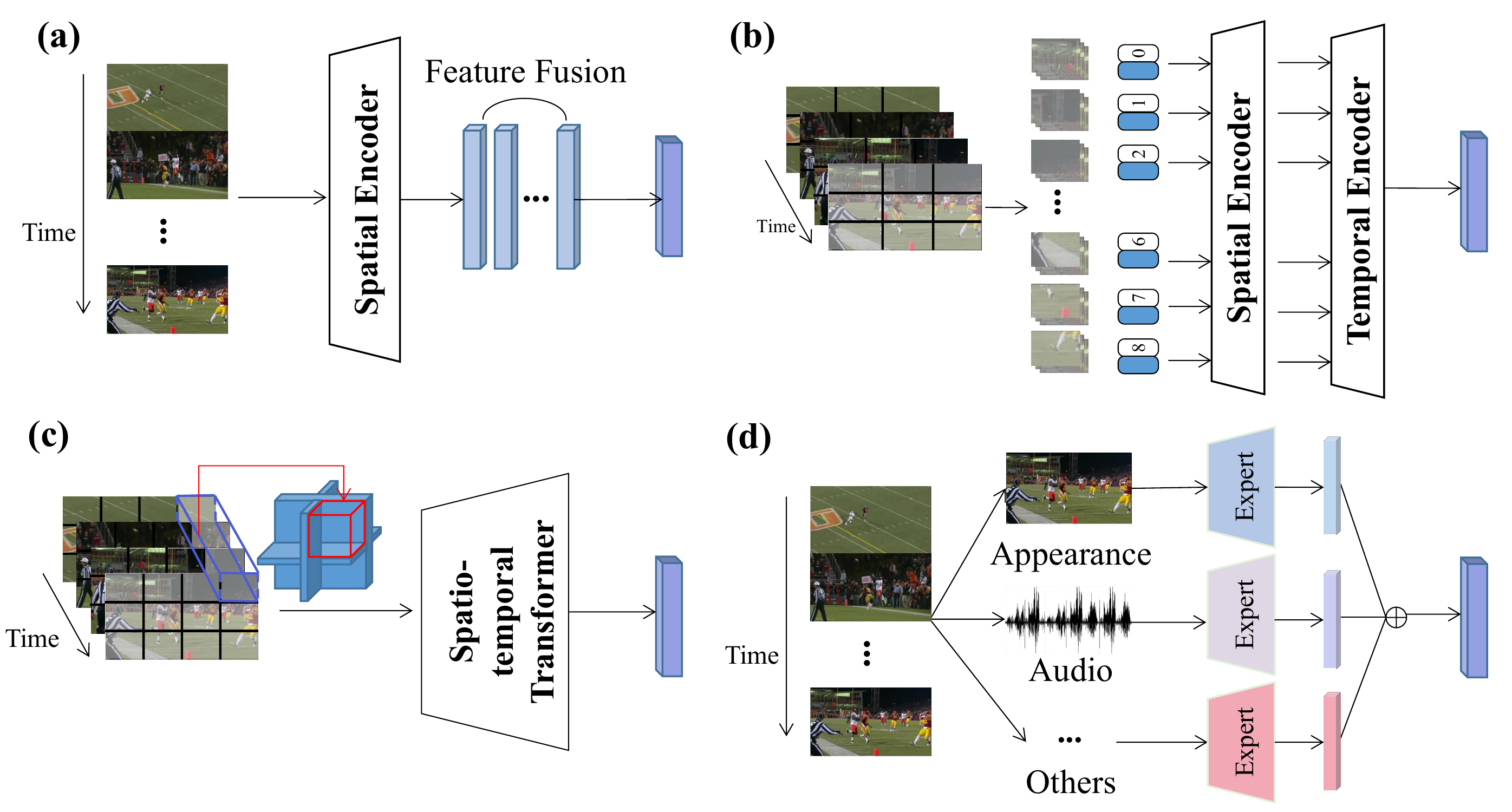}
\caption{For a given video, several key frames are first obtained by sampling. Due to its spatio-temporal characteristics, the process in (a) and (b) is to employ CNNs and Transformer to extract spatial features and then utilize RNNs, CNNs, Transformer, or other feature fusion architectures to extract temporal features. In (c), it directly employs the spatio-temporal Transformer to obtain video representation. Since videos contain multiple modalities, more experts can be selected to perform the extraction of each modality information and are concatenated as the final representation as shown in (d).
}
\label{Video Feature Extraction fig}
\end{figure*}

\section{Video Representation Extraction}\label{Video Represetation}
Video representation extraction specifies a way of representing a video. Compared to static image data, video data is a type of information-intensive and complex media, which contains both spatial and temporal representation. Current studies~\cite{dong2019dual,bain2021frozen,luo2021CLIP4clip} are dedicated to learning both spatial and temporal representations. Besides, since the multiple modalities contained in videos, several works devote to multi-modal feature extraction. With the rapid progress of computing resources and the availability of large-scale data, deep learning has become a popular way to capture features. We will detail deep learning approaches for video representation extraction in terms of spatial feature (Section~\ref{Video_spatial}), temporal feature (Section~\ref{Video_temporal}) and multi-modal feature (Section~\ref{video_expert}), respectively. The general framework for video representation extraction is illustrated in Fig.~\ref{Video Feature Extraction fig}.
In addition, we categorize the video feature extraction methods in Table~\ref{classify}, where the first column lists the approaches that employ multiple expert models, and the second and third columns show the basic networks to extract temporal and spatial representation separately.

\subsection{Spatial Feature Extraction}\label{Video_spatial}
In general, the first step for representing a given video is to select the key frames. The most common approaches for sampling video frames include
random sampling (\ie, sampling several frames per second)~\cite{miech2018learning,sun2019videobert,ging2020coot,ge2022miles}, uniformly sampling a fixed number of frames~\cite{yang2020tree,cheng2021improving}, and sparse sampling~\cite{lei2021less}. 
Afterward, the next step is to extract features for the selected frames. Since each frame in a video is similar to a still image, its spatial feature can be implemented similarly to the still image.  Previous methods~\cite{miech2018learning,mithun2018learning,yu2018joint,dong2019dual} are mainly based on CNNs variants, while the Transformer architecture has become widely used since 2018. Overall, we divide existing methods into CNNs-based and Transformer-based spatial feature extractors.

\textbf{CNNs based methods.} Many popular CNNs architectures, including AlexNet~\cite{krizhevsky2012imagenet}, VGGNet~\cite{simonyan2014very}, GoogLeNet~\cite{szegedy2015going}, ResNet~\cite{he2016deep} and DenseNet~\cite{huang2017densely}, have been applicable in extracting the features from sampled video frames.
For instance,~Torabi et al.~\cite{torabi2016learning} apply a pre-trained VGG-19 model to extract frame-level features from the penultimate layer. Mithun et al.~\cite{mithun2018learning} resize frames and input them into ResNet-152 to capture frame-wise features from the penultimate layer.
Likewise, many works~\cite{mithun2018learning,yang2020tree,liu2021hit} extract video appearance features from the global average pooling in a pre-trained ResNet152 or SENet-154~\cite{hu2018squeeze}.~Wang et al.~\cite{wang2022cross} extract frame features from the FC layer whose input is the last average pooling layer in ResNet-152.

\textbf{Transformer based methods.} Transformer~\cite{vaswani2017attention} has developed rapidly and achieved remarkable progress in recent years, which abandons primary neural networks and solely contains stacked encoder-decoder blocks composed of multi-head self-attention (MSA), Multi-Layer Perceptron (MLP), and layer-norm (LN). For instance, VSR-Net~\cite{han2021visual} employs a multi-layer Spatial Transformer block to perform interactions among objects and then adopts average pooling to generate frame representation. Besides,~Dosovitskiy et al.~\cite{dosovitskiy2020image} provide a general Vision Transformer (ViT) for extracting image or video frame features. ViT is proposed as a pure transformer architecture for image classification, which generates excellent performances and becomes a strong backbone already~\cite{cheng2021improving,han2021visual,ge2022miles}.
To be specific, the first step is to convert a given sampled frame into flattened and discrete $2D$ patches. Then, it projects those non-overlapping patches into $1D$ tokens via linear projection, meanwhile, recording their position and classification information via position and extra learnable $[class]$ embeddings. After feeding them into a Transformer structure, the frame representation is generated from the last layer of the $[CLS]$ token. For instance, CLIP4Clip~\cite{luo2021CLIP4clip} introduces a 3D linear to learn temporal information across frames.
Notably, ViT has multiple variants, \eg, ViT B/16, ViT L/14, ViT H/14, and the selection of different models may affect the final results (See Tab.~\ref{finetune evaluation}).
Moreover, as the proposal of large-scale pre-training methods, CLIP~\cite{radford2021learning} provides a pre-trained ViT encoder to directly extract vision features. 
In one word, Transformer has gradually become the mainstream network architecture for VTR~\cite{gao2021clip2tv,fang2021clip2video,gorti2022x,ma2022x}, due to its excellent performance on a variety of public datasets.

\subsection{Temporal Feature Extraction}\label{Video_temporal}
After extracting spatial features, the next step is to model their temporal interaction information.
One simple and general manner is aggregating features of the sampled frames via mean pooling~\cite{mithun2018learning,dong2018predicting,dong2019dual,liu2021hit} or max pooling~\cite{miech2019howto100m,wray2019fine}. In addition, temporal features in videos can be modeled by RNNs and CNNs. Moreover, recent state-of-the-art methods exploit Transformer to generate more sophisticated temporal features~\cite{sun2019videobert,tan2019lxmert,su2019vl}. 

\textbf{RNNs based methods.} 
Motivated by the significant performance of RNNs in the field of NLP~\cite{bengio1994learning}, some studies \cite{dong2019dual,ali2022video} transfer those typical networks, especially GRU networks, to capture long-range temporal features in videos. Concretely,~Dong et al.~\cite{dong2022reading} input frame features into bidirectional GRU (biGRU)~\cite{chung2014empirical} to extract temporal information from both forward and backward and then perform mean pooling to generate the whole representation along the temporal dimension. After training biGRU, Dong et al.~\cite{dong2019dual} input the feature maps into 1D CNNs, which includes several filter sizes to capture multi-scale features. Besides, Yang et al.~\cite{yang2020tree} capture temporal features via transforming spatial features extracted from ResNet152~\cite{he2016deep} to GRU and then aggregate them via an attention module. Despite the fact that RNNs have been applied in many studies, one notable weakness is its expensive training time, in particular for very long-range videos.

\textbf{CNNs based methods} also achieve excellent results for VTR in these years. In terms of 2D CNNs, the work in~\cite{lin2019tsm} has designed a temporal shift module (TSM) which is inserted into the residual branch in ResNet-50 for shifting some channels between frames in the temporal dimension and fusing temporal information among multiple frames. SSAN~\cite{guo2021ssan} demonstrates that learning spatial correlations firstly can provide extra information for better temporal correlations extraction. Compared with 2D CNNs, 3D CNNs add a temporal dimension during moving across the channels, which is favorable for capturing spatio-temporal features. To this end, many approaches directly extend 2D CNNs into 3D CNNs instead~\cite{tran2015learning,carreira2017quo,tran2017convnet}.
For instance, Mithun et al.~\cite{mithun2018learning} extract activity (motion) features via I3D~\cite{carreira2017quo} which inflates 2D CNNs to deep 3D CNNs. Res3D~\cite{tran2017convnet} extends all $3\times 3$ convolutional layers in ResNet~\cite{he2016deep} by $3\times 3\times 3$, which is applied in SlowFast model~\cite{feichtenhofer2019slowfast} that concatenates spatial and temporal features in two branches from dense frames and fewer channels. Besides,~Feichtenhofer et al.~\cite{feichtenhofer2016convolutional} demonstrate that fusing spatial features in the last layer with 3D CNNs and replacing 2D pooling with 3D pooling can learn better representation. Substitution of the first three layers also improves performance, as confirmed in~\cite{karpathy2014large}.
Furthermore, Hara et al.~\cite{hara2018can} show the effectiveness of building deeper 3D CNNs for large-scale video datasets. 

While 3D CNNs bring the above advantages, the retrieval speed, as a significant evaluation metric, has slowed down significantly. To overcome this deficiency, the pseudo-3D CNNs model is proposed~\cite{sun2015human,qiu2017learning,tran2018closer}, which replaces 3D CNNs with spatial 2D CNNs and temporal 1D CNNs. Likewise, Separable 3D (S3D)~\cite{xie2018rethinking} is to substitute 3D CNNs in I3D~\cite{carreira2017quo}, which is commonly utilized for extracting motion features~\cite{wang2021t2vlad,gabeur2020multi,liu2019use,sun2019learning,sun2019videobert}. Additionally, they add gating mechanisms after each 1D CNNs of S3D to generate S3D-G for further improving accuracy. 
For considering fine-grained features, Local and Global Diffusion (LGD)~\cite{qiu2019learning} is presented to strengthen interactions between local and global feature representation via combining LGD-2D~\cite{yao2018yh} and LGD-3D (a pseudo-3D CNNs~\cite{qiu2017learning}). 
In a nutshell, applying 2D CNNs to extract spatial features and then utilizing 3D CNNs to capture temporal motion features is a common way for video feature extraction based on CNNs~\cite{mithun2018learning,han2021visual}.

\textbf{Transformer based methods.} In the previous subsection, we have introduced the application of Transformer in spatial feature extraction. In fact, Transformer also has a great ability to capture long-distance temporal relations. a few works~\cite{gao2021clip2tv,han2021visual,ma2022x,wang2022disentangled} propose to build Transformer based models to capture the interaction information between different frames. For example, CLIP2Video~\cite{fang2021clip2video} proposes to model the differences (\ie, motion features) among consecutive adjacent frames and feed them into a temporal Transformer with position and type information to generate temporal features via average pooling. COOT~\cite{ging2020coot} introduces a temporal Transformer to capture frame and clip feature interactions successively. X-CLIP~\cite{ma2022x} applies a three-layer Transformer to encode frame features and averages them to obtain video features. Han et al.\cite{han2021visual} utilize a multi-layer Transformer to capture spatial features among adjacent frames, and an attention-aware feature aggregation layer to fuse features into a comprehensive representation. In addition, CLIP~\cite{radford2021learning} also provides four layers of temporal transformer blocks (with frame position embedding and residual connection), which is widely employed in numerous works~\cite{gao2021clip2tv,wang2022disentangled,min2022hunyuan_tvr}.

Recently, because of the excellent performance of ViT in capturing spatial representation, many researches~\cite{bertasius2021space,zhang2021vidtr,arnab2021vivit,bain2021frozen} are dedicated to developing novel Transformer networks for learning both spatial and temporal features. Not like the mentioned methods that capture spatial features first and then transfer them into a temporal Transformer, Frozen~\cite{bain2021frozen} proposes a stack of space-time Transformer blocks, which can learn both temporal and spatial position, feed flattened spatial-temporal patch embeddings into temporal and spatial self-attention layers sequentially. 
Likewise, TimeSformer~\cite{bertasius2021space} introduces a richer study of five spatio-temporal combinations, indicating that the divided space-time scheme achieves the best performance because of its larger training capacity, especially when performing on longer video clips and higher frame resolution. Recent work in~Ge et al.~\cite{ge2022miles} applies 12 divided space-time self-attention blocks to obtain feature representation, utilizing masked visual modeling to mask out the local content of consecutive frames and obtain more fine-grained video information.

\subsection{Multi-modal Video Feature Extraction}\label{video_expert}
Since the video contains not only spatio-temporal characteristics but also multi-modal information, such as audio, Optical Character Recognition (OCR), and motion. Some works also study how to extract multi-modal video features. For these modalities, additional `Experts' are integrated, each of which is dedicated to extracting features from a specified modality and can be applied directly~\cite{miech2018learning,liu2019use,gabeur2020multi,wang2021t2vlad}. Existing studies~\cite{dzabraev2021mdmmt,kunitsyn2022mdmmt} combine several modality information with spatial-temporal features and demonstrate the ability to enrich final semantic representation. The feature extraction methods for each expert are introduced below. 

\begin{itemize}
    \item \textbf{Scene} embeddings are extracted from DenseNet-161~\cite{huang2017densely} trained for image classification on the Places365 dataset~\cite{zhou2017places}. 
    \item \textbf{Face} features are extracted in two stages: firstly extract bounding boxes via an SSD~\cite{liu2016ssd} face detector, and then pass into ResNet50 that trained for face classification on the VGGFace2 dataset~\cite{cao2018vggface2}.
    \item \textbf{Motion} features are extracted from S3D~\cite{xie2018rethinking} and SlowFast~\cite{feichtenhofer2019slowfast} trained on Kinetics action recognition dataset, or I3D~\cite{carreira2017quo} and a 34-layer R(2+1)D~\cite{tran2018closer} trained on IG-65m.
    \item \textbf{Audio} features are extracted based on VGGish~\cite{hershey2017CNN} trained on YouTube-8m dataset for audio classification.
    \item \textbf{OCR} features are obtained in two steps: the overlaid text is first detected via the pixel link text detection model~\cite{deng2018pixellink}. Then the detected boxes are passed through a text recognition model trained on the Synth90K dataset.
    \item \textbf{Speech} transcripts are extracted using the Google Cloud Speech to Text API, with the language set to English. The detected words are then encoded by pre-training word2vec\cite{mikolov2013efficient}. 
\end{itemize}

After extracting video features of each modality, the primary question is how to perform temporal aggregation to generate video representations composed of multiple expert features. Yu et al.~\cite{yu2018joint} reduce the dimension of extracted audio features and directly concatenate them with visual features to obtain the final video representation. As suggested by~Miech et al.~\cite{miech2018learning} and Liu et al.~\cite{liu2019use}, the gating mechanism enables to act as an effective way to predict and choose attention relationships among experts and further to fuse them via average pooling or the VLAD mechanism. 
Moreover, Shvetsova et al.~\cite{shvetsova2021everything} raise the Multi-modal Fusion Transformer to fuse multi-modal features and realize modality agnostic via training all combinations among different modalities, by averaging them and projecting them together with normalization.

\begin{figure*}[hbt!]
\centering
\includegraphics[width=\textwidth]{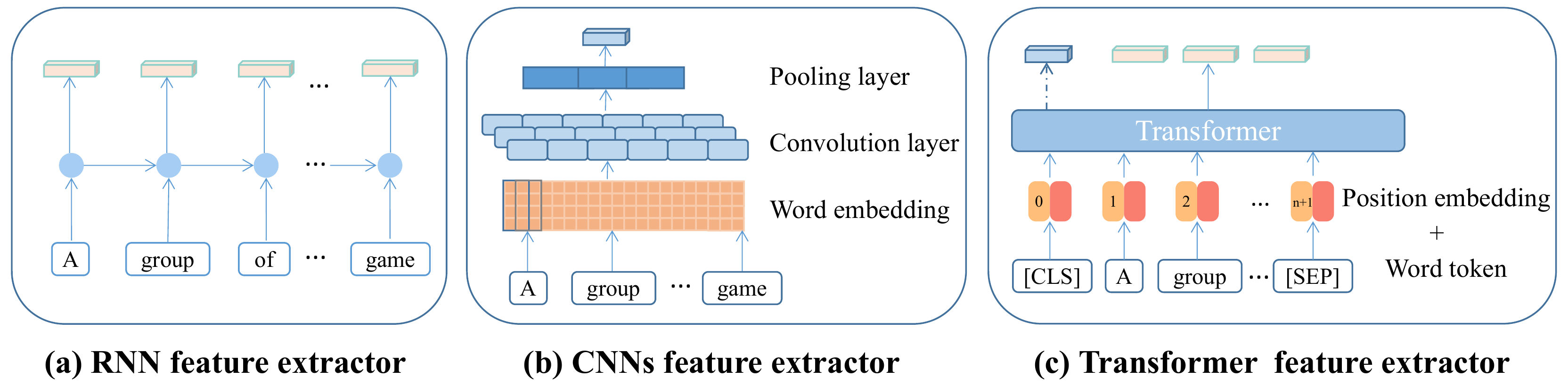}
\caption{Three networks for textual representation generation: (a) RNNs based textual feature extraction; (b) CNNs based textual feature extraction; (c) Transformer based textual feature extraction.}
\label{Text Feature Extraction}
\end{figure*}

\textbf{Takeaway for video representation.} 
In this section, we have summarized typical approaches for capturing video representation. 
Based on the spatio-temporal characteristics within videos, the temporal and spatial features are extracted respectively. CNNs, Transformer, and even RNNs based architectures can be chosen to generate comprehensive video representation for processing frame-wise features, or dividing them into several hierarchies for fine-grained feature interaction. Additionally, more `experts' can be directly utilized to extract more multimedia information. Due to the relatively high performance achieved by CLIP model so far, researchers strike to build appropriate networks according to different task scenarios to extract the most important modality information and generate a more comprehensive video representation.

\section{Textual Representation Extraction}\label{Text Representation}
Textual representation extraction aims to extract features from language sentences. The main challenge is how to model sequential relationships to capture complete semantic information. Since the sophisticated developments of NLP with deep neural networks, the capacity for processing long-term dependencies problems is continuously strengthened. At the early stage, word2vec and Glove are adopted to extract word embeddings. Later, RNNs with strong long-distance learning ability have been adopted in more studies~\cite{otani2016learning,yang2020tree}. In some studies, CNNs may achieve semantic representation ability similar to RNNs. Since the recent emergence of Transformer variants, the problem that RNNs cannot operate in parallel has been solved, and a breakthrough in performance has been achieved on many benchmarks. The following sections aim to survey the approaches of textual representation based on RNNs (Section~\ref{text-RNNs}), CNNs (Section~\ref{text-CNNs}), and Transformer (Section~\ref{text-Transformer}). The common strategies for textual representation generation are shown in Fig.~\ref{Text Feature Extraction}. Meanwhile, Table~\ref{classify} summarizes the use of different textual feature extractors. 

\subsection{RNNs based Textual Feature}\label{text-RNNs}
Recurrent Neural Networks (RNNs)~\cite{bengio1994learning} are proposed to learn associations between words in the sentences for modeling long sequence data, while the problems of gradient disappearance and gradient explosion arise. To solve the problem, LSTM represents long short-term memory~\cite{hochreiter1997long}, which aims to make up for the problems caused by RNNs via remembering something crucial and choosing unimportant to forget in long sequence data. For instance, to address the problem of long sequence forgetting, Otani et al.~\cite{otani2016learning} utilize RNNs to extract textual features for generating sentence embedding. 
Bi-directional LSTM (biLSTM) is employed twice in work~\cite{chen2020fine} to obtain contextual-aware word embeddings on graph nodes that are distinguished and generated by off-the-shelf semantic role parsing toolkit~\cite{song2019polysemous}.

GRU~\cite{chung2014empirical} stands for Gated Recurrent Unit, which is a modification based on LSTM. The input gate and the forget gate are combined to generate the update gate, and the cell state and hidden unit are fused as well. Compared with LSTM, the GRU model is simplified with less calculation. The original sentences are input into GRU via the word embedding transformation to obtain text representation~\cite{mithun2018learning}. HANet~\cite{wu2021hanet} utilizes biGRU to select features corresponding to verbs and nouns as individual-level representations. The modified relational GCN~\cite{schlichtkrull2018modeling} is then employed to obtain local and global-level representation. Besides, Tree-structured LSTM~\cite{tai2015improved} is proposed to capture semantic features based on word nodes relationship of tree structure~\cite{yang2020tree}.

\subsection{CNNs based Textual Feature}\label{text-CNNs}
Since 2014, Yoon Kim and other researchers~\cite{zhang2015sensitivity,chen2015convolutional} have introduced CNNs to the NLP domain for extracting better short-range text features with fast calculation, including the related input layer, convolution layer, pooling layer, and fully connected layer. MSSP~\cite{fan2020person} is proposed to recognize the specific person via natural language with bounding boxes, which utilizes 1D CNNs to encode sentences and then captures local features of sentences across multiple scales. Besides, they also choose GRU and Gaussian-Laplacian mixture models to improve the final performance. Miech et al.~\cite{miech2018learning} apply the aggregation module of NetVLAD~\cite{arandjelovic2016netvlad} to aggregate the input word embedding vectors to a global vector representation. Also, caption features are generated from a shallow 1D-CNNs which is built on the top of pre-computed word embeddings~\cite{miech2019howto100m}.

\subsection{Transformer based Textual Feature}\label{text-Transformer}
With the emergence and development of Transformer, it almost replaces RNNs in the NLP field gradually. As BERT is put forward, the development of the NLP field goes to a higher level, and it might be the first choice to extract textual features~\cite{gabeur2020multi,ging2020coot,liu2021hit,wang2021t2vlad}. 

BERT is short for Bidirectional Encoder Representations from Transformer~\cite{devlin2018bert}, which includes multiple stacked combinations of Multi-Head Attention, Add \& Norm, Feed Forward, and Residual Connection. The sum of token embeddings, segment embeddings, and position embeddings are fed into BERT for learning global semantic information. Token $[CLS]$ situates before the first token of the sentence for text classification, and $[SEP]$ is for separating two sentences. The $[CLS]$ token from the last layer often represents global semantic information. For pre-training, the Masked Language Model (MLM) and Next Sentence Prediction (NSP) are applied to capture more comprehensive word and sentence-level representations, respectively. The core of MLM is masking several words in each sentence randomly and then predicting them by the remaining words. Given two sentences in an article, NSP is applied to determine whether the second sentence follows the first one in the text. BERT has become the most common architecture for the VTR task in recent years. Since the different numbers of encoders and hidden layers, BERT can be divided into BERT-Tiny, BERT Mini, BERT-Small, BERT- Medium, and BERT-Base. BERT-Base is the most commonly applied encoder in many works~\cite{gabeur2020multi,ging2020coot,liu2021hit,wang2021t2vlad}. Besides, due to the differences in case sensitivity, there are BERT-uncased~\cite{ging2020coot,liu2021hit} and BERT-cased~\cite{wang2021t2vlad}. 

Moreover, due to the excellent performance achieved by BERT, plenty of BERT-like architectures are proposed successively, \eg, RoBERTa~\cite{liu2019roberta}, ALBERT~\cite{lan2019albert} and DistilBERT~\cite{sanh2019distilbert}. Particularly, DistilBERT~\cite{sanh2019distilbert} introduces knowledge distillation to reduce the model size and inference speed without excessively reducing the performance, which is applied for video-text retrieval tasks such as in Frozen~\cite{bain2021frozen} and MILES~\cite{ge2022miles}.
More recently, CLIP~\cite{radford2021learning} provides a Transformer architecture for extracting textual features and achieves state-of-the-art performance in extensive experiments~\cite{gao2021clip2tv,luo2021CLIP4clip,fang2021clip2video,zhao2022centerclip}.

\textbf{Takeaway for textual representation.} 
First, the application of RNNs solves the problem of long-range feature extraction to a certain extent. While the problems of gradient disappearance and gradient explosion are alleviated in subsequent LSTM and GRU, these models do not process parallel computing due to long-time contextual dependency.
Second, CNNs have strong parallel computing capability, but compared with RNNs, they cannot model long-distance features. It can be seen that the number of convolution layers may increase to compensate for this deficiency.
Third, compared with RNNs and CNNs, Transformer is currently more widely utilized and its capability of long-term feature extraction is slightly better than RNNs. Besides, Transformer is better in terms of the overall ability to extract text features. The work in~\cite{radford2021learning} compares several typical textual feature extractors.

\section{Feature Embedding and Matching}\label{Feature matching}
\begin{figure*}[hbt!]
\centering
\includegraphics[width=\textwidth]{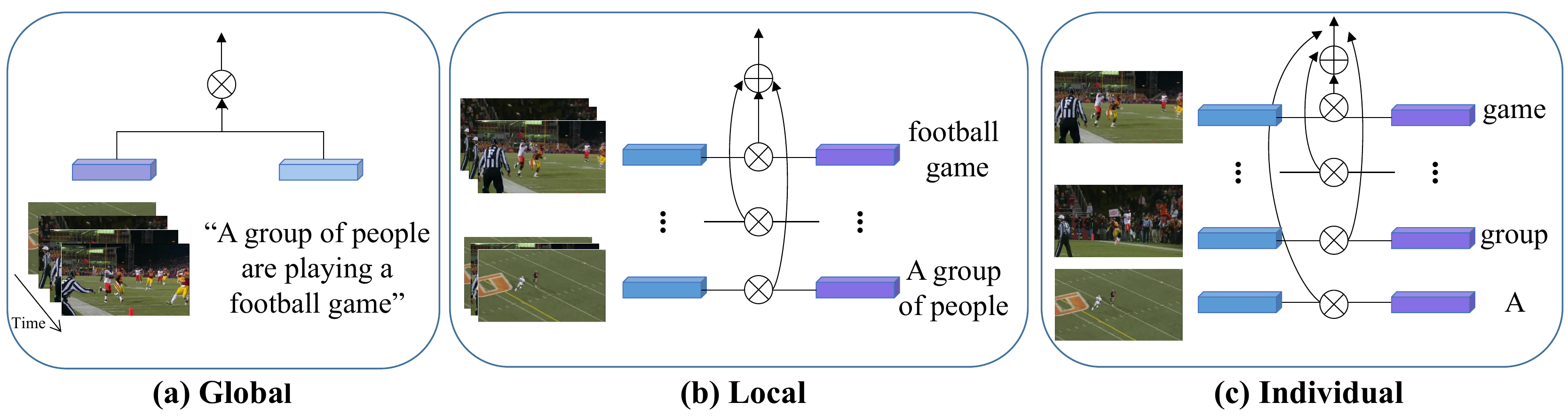}
\caption{Three feature matching strategies: (a) feature matching on the global level; (b) feature matching on the local level; (c) feature matching on the individual level. Note that, global is video-sentence level, local is clip-phrase level, and individual is matching on the frame-word level.}
\label{Similarity}
\end{figure*}

After generating video and textual feature representation, the key is how to project them into a joint embedding space for feature alignment and similarity calculation. Most existing works~\cite{yang2020tree,luo2020univl} only align the global cross-modal features without considering local details. With the gradual mining of hierarchical semantics, some studies~\cite{chen2020fine,ging2020coot} divide features into more fine-grained hierarchical, \ie, local features or even individual features. Finally, combining each level alignment improves the feature matching accuracy. Figure~\ref{Similarity} shows the process of video-text matching methods from three categories: Global, Local, and Individual, which are detailed in Section~\ref{Feature matching with global}, Section~\ref{Feature matching with local} and Section~\ref{Feature matching with individual}, respectively. Additionally, Fig.~\ref{1} shows the combinations of different granularity matching and lists the representative works for each combination in recent years.

\subsection{Global Matching}\label{Feature matching with global}
\begin{figure}[hbt!]
\centering
\includegraphics[width=0.5\textwidth]{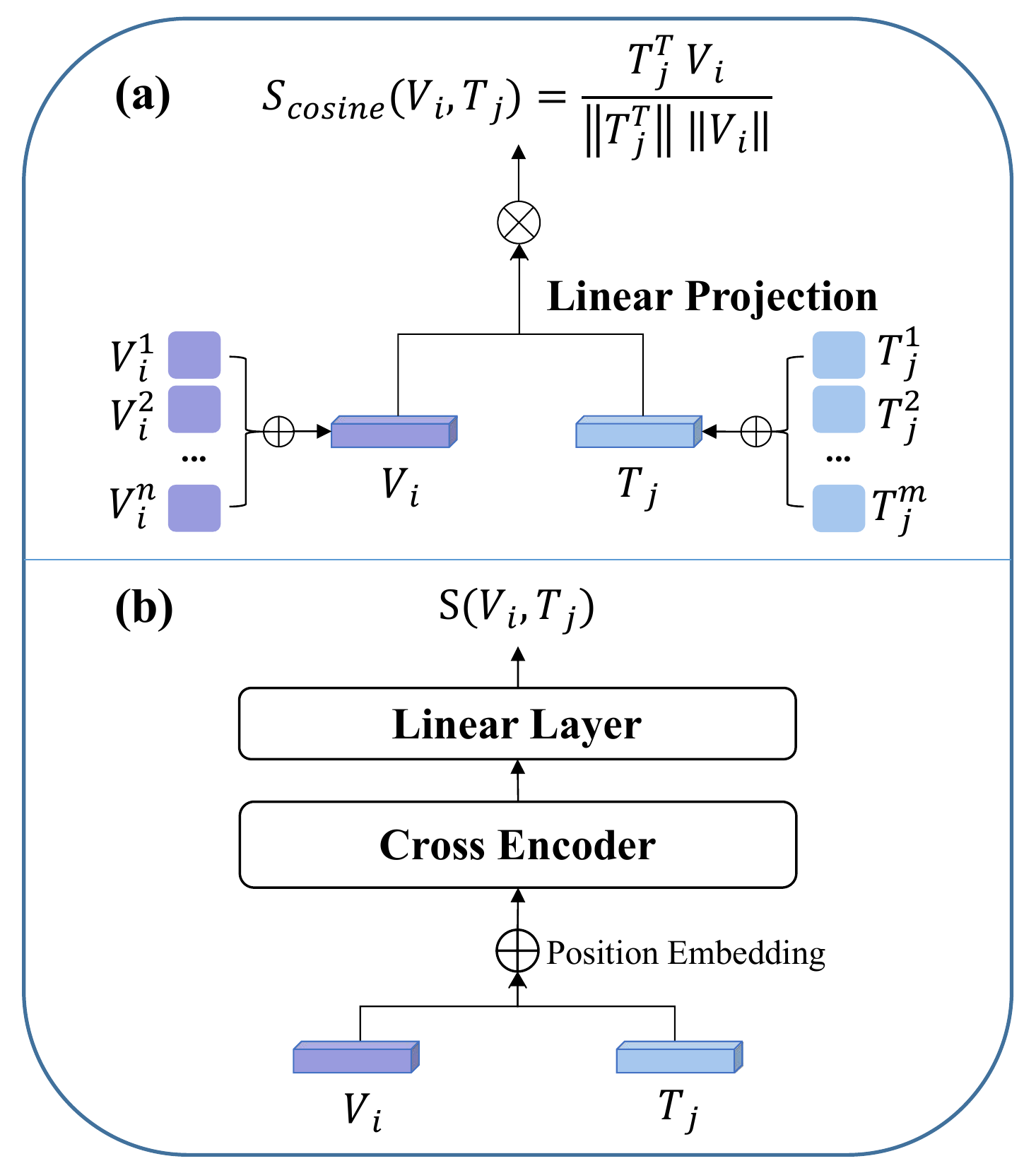}
\caption{Two global matching strategies: (a) coarse-grained feature embedding via linear projection and matching through $cosine$ function; (b) multi-modal features interact in a cross encoder and generate similarity scores via a linear layer.}
\label{global matching}
\end{figure}

Global matching aims to project coarse-grained video and text features into a joint embedding space and then perform global similarity computation. Two categories of global matching are depicted in Fig.~\ref{global matching}. We take the linear projection and $cosine$ similarity calculation as an example to present the most commonly applied way in Fig.~\ref{global matching}(a). Fig.~\ref{global matching} (b) demonstrates the way to capture video and text feature correlations through Transformer and output the predicted score via a linear layer. Next, we detail how to project and calculate these similarities.

Linear projection is the easiest way to map features into low dimension~\cite{yang2020tree,ging2020coot,bain2021frozen}. For instance, ActBERT~\cite{zhu2020actbert} utilizes a linear layer on both ``$[CLS]$'' tokens of video and text and a $sigmoid$ function to obtain similarity scores. VSR-Net~\cite{han2021visual} adopts a pointwise linear layer and an attention-aware feature aggregation layer to generate video and text embedding vectors for latter matching. Furthermore, linear projection matrices are adopted in works~\cite{yang2020tree,ali2022video}, and X-Pool~\cite{gorti2022x} performs projection with a Layer Normalization layer and a projection matrix. Additionally, non-linear embedding functions are usually implemented by means of FC layers~\cite{dong2022reading} and gating functions~\cite{miech2018learning,miech2019howto100m}. Several studies~\cite{lu2016event,dong2019dual} apply VSE++~\cite{faghri2017vse++} for projecting features into a joint embedding space with affine
transformation. Besides, MLPs can also be adopted as the nonlinear projection heads to conduct nonlinear transformations~\cite{liu2021hit}. Additionally, Cross encoder is introduced to concatenate video and text features in the temporal dimension with their position embeddings for capturing multi-modal interaction~\cite{luo2020univl,luo2021CLIP4clip}.

The ways for computing matching scores contain dot-product~\cite{bain2021frozen}, cosine distance, Jaccard function~\cite{wu2021hanet}, and Euclidean distance. Cosine distance is adopted in most works~\cite{liu2019use,miech2019howto100m,dong2019dual,wu2021hanet}. Moreover, a few works~\cite{ma2022x,ali2022video} adopt matrix multiplication between video and text representations to assess similarities. Take CLIP4Clip~\cite{luo2021CLIP4clip} as an example, it divides the similarity calculator into three categories based on whether new parameters are introduced. In this survey, we determine the category of feature matching according to the extracted categories of features used to calculate similarity. Therefore, we consider it as global matching, which generates the global representation by mean
pooling and applies cosine similarity to calculate similarity scores for the first two calculators. Additionally, the third ``Tight type'' realizes cross-modal feature interaction with a Transformer encoder and then generates the prediction score via two linear layers and an activation function.

\subsection{Local Matching}\label{Feature matching with local}
Different from global matching, local matching estimates whether the local features (\ie, words and frames) can be aligned or not. Especially, some studies demonstrate that the retrieval performance of global matching is even lower than that of only local matching. However, combing the two strategies can achieve remarkable performance boosts~\cite{wang2021t2vlad,liu2021hit,han2021visual,ma2022x}.

To perform local matching, in addition to projection methods described previously, several feature aggregation and embedding strategies are adopted on word and frame features. Clustering center is a common way, which contains k-means~\cite{sun2019videobert}, NetVLAD~\cite{arandjelovic2016netvlad} and GhostVLAD~\cite{zhong2018ghostvlad}. Generally, local feature embeddings are projected into multiple cluster centers according to soft-aligned weights, which are usually calculated by applying the dot product between features and cluster centers. Those centers are then aggregated and normalized as fine-grained embeddings. At last, similarity calculation is performed on these fine-grained embeddings between video and text to generate local matching scores~\cite{wang2021t2vlad,wang2022hybrid}. For instance, CenterCLIP~\cite{zhao2022centerclip} arises a multi-segment clustering strategy (\ie, k-medoids++ and Spectral clustering) for capturing detailed temporal interaction information and enhancing feature alignment on segment levels. Besides, since experts are utilized to extract multi-modal information in the video, for achieving video-text alignment, most studies~\cite{liu2019use,gabeur2020multi} apply NetVLAD to cluster text features and project them into separate subspaces according to each expert. Considering the different importance of multi-modal features for VTR, a weighted sum of cosine distances can be calculated for different experts~\cite{miech2018learning,wang2021t2vlad}.

In addition to projecting text and video features into a single joint space, other multiple embedding spaces are also exploited to perform alignment and similarity calculations. Mithun et al.~\cite{mithun2018learning} map appearance, audio, and motion features into Object-Text and Activity-Text space separately by multiplying transformation matrices. 
VSR-Net~\cite{han2021visual} projects text, video features, and relational features into relation and video embedding, and the final similarity is then calculated by multiple embedding spaces respectively. \textcolor{black}{CAMoE~\cite{cheng2021improving} computes similarities from three aspects: entity, sentence, and action.}

Like global matching, many approaches apply cosine similarity to estimate local matching~\cite{mithun2018learning,cheng2021improving,han2021visual}. In addition, HiT~\cite{liu2021hit} proposes local features with semantic and feature levels and aligns both positive and negative similarities on these two levels. Besides, a variety of attention mechanisms are used to align each level of features and normalize similarity weights via $softmax$ function~\cite{chen2020fine,ging2020coot,wu2021hanet}. For instance, HANet~\cite{wu2021hanet} utilizes an attention mechanism to enhance local-level cosine similarity and the Jaccard similarity on the concept level. Moreover, matrix multiplication is adopted in work~\cite{ma2022x} to compute cross-grained features similarities. 
At last, since multiple feature similarity scores of different levels are generated, it is straightforward to sum and average them to represent the final similarity score~\cite{chen2020fine,wu2021hanet,wang2021t2vlad}.

\subsection{Individual Matching}\label{Feature matching with individual}
Some studies take into account not only global and local feature matching but also a more fine-grained feature hierarchy, \ie, performing feature alignment at the individual level. Generally, several works adopt matrix multiplying to calculate similarities~\cite{ma2022x,min2022hunyuan_tvr}. To be specific, Min et al.~\cite{min2022hunyuan_tvr} extract features with the same dimension and then compute each frame-word pair similarity via dot-product to form a similarity matrix, which is then aggregated to generate the individual similarity score.~Wang et al.~\cite{wang2022disentangled} add weights on relational individual pairs via the attention module. HANet~\cite{wu2021hanet} utilizes an attention mechanism to calculate the weighted cosine individual similarity.

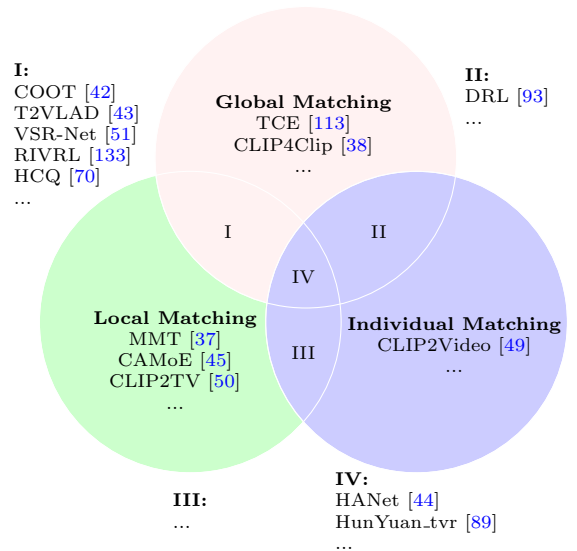
\begin{figure}[!htbp]
\centering
\footnotesize
\begin{tikzpicture}
	\draw (0,0)[color=white,domain=0:2.0,fill=green!20] circle (2cm);
	\draw (55:2.67cm)[color=white,domain=0:2.0,fill=pink!20] circle (2cm);
	\draw (0:3cm)[color=white,domain=0:2.0,fill=blue!20] circle (2cm);
	\draw (0,0)[color=white,domain=0:2.0] circle (2cm);
	\draw (55:2.67cm)[color=white,domain=0:2.0] circle (2cm);
	\draw (0:3cm)[color=white,domain=0:2.0] circle (2cm);
	\node at (1.5,2.5)[align=center]{\bf Global Matching\\TCE~\cite{yang2020tree}\\CLIP4Clip~\cite{luo2021CLIP4clip}\\...};
	\node at (-0.2,-0.5)[align=center] {\bf Local Matching\\MMT~\cite{gabeur2020multi}\\CAMoE~\cite{cheng2021improving}\\CLIP2TV~\cite{gao2021clip2tv}\\...};
	\node at (3.5,-0.3)[align=center] {\bf Individual Matching\\CLIP2Video~\cite{fang2021clip2video}\\...};
	\node at (0.5,1.2) {\uppercase\expandafter{\romannumeral1}};
	\node at (2.5,1.2) {\uppercase\expandafter{\romannumeral2}};
	\node at (1.5,-0.4) {\uppercase\expandafter{\romannumeral3}};
	\node at (1.5,0.6) {\uppercase\expandafter{\romannumeral4}};
	
	\node at (-1.5,2.5)[align=left] {\bf \uppercase\expandafter{\romannumeral1}:\\COOT~\cite{ging2020coot}\\T2VLAD~\cite{wang2021t2vlad}\\VSR-Net~\cite{han2021visual}\\RIVRL~\cite{dong2022reading}\\HCQ~\cite{wang2022hybrid}\\...};
	\node at (4.2,3)[align=left] {\bf \uppercase\expandafter{\romannumeral2}:\\DRL~\cite{wang2022disentangled}\\...};
	\node at (0,-2.5)[align=left] {\bf \uppercase\expandafter{\romannumeral3}:\\...};
	\node at (3,-2.5)[align=left] {\bf \uppercase\expandafter{\romannumeral4}:\\HANet~\cite{wu2021hanet}\\HunYuan\_tvr~\cite{min2022hunyuan_tvr}\\...};
\end{tikzpicture}
\caption{Combinations of three matching strategies.}\label{1}
\end{figure}

\textbf{Takeaway for feature embedding and matching.} Feature embedding and matching is to map video and textual features into a joint embedding space via linear, nonlinear projection, or clustering centers. 
According to the fine-grained features, different levels of matching strategies are developed. The final similarity result is obtained by summing and averaging the similarity score from different granularity levels. As can be seen in Fig.~\ref{1}, more works combine global and local matching, while none of them combine local and individual matching without using global matching yet.

\section{Objective Functions}\label{Training Objectives}
This section introduces objective functions for optimizing and constraining feature representation and matching.
Metric learning methods, which include triplet loss and contrastive loss, are mostly utilized to reduce the intra-class distance and increase the inter-class distance. Since VTR is a bidirectional retrieval task, most loss functions are summed in two directions, \ie,
$ \mathcal{L}_{total}=\left(\mathcal{L}_{t 2 v}+\mathcal{L}_{v 2 t}\right)$.
We will detail several typical loss functions below, including triplet loss functions in Sec.~\ref{Triplet Ranking Loss}, contrastive loss functions in Sec~\ref{Contrastive Loss}, and other loss functions~\ref{Other loss}.

\subsection{Triplet Ranking Loss Functions}\label{Triplet Ranking Loss}
Ranking loss is to evaluate and modify the similarity scores by measuring the distance between samples. 
It is firstly proposed to verify face embedding similarities between the anchor-positive and anchor-negative pairs~\cite{schroff2015facenet}.
Specific to VTR, the input triplet consists of a positive video-text sample pair and a negative video or text sample. The loss function aims to maximize the distance of unmatched video-text pairs and minimize that of the matched pairs. The commonly adopted triplet ranking loss includes Bi-directional max-margin ranking loss~\cite{yu2016video,miech2018learning,miech2019howto100m,liu2019use,gabeur2020multi,wang2021t2vlad,kunitsyn2022mdmmt} and Hinge-based triplet ranking loss~\cite{faghri2017vse++,patrick2020support,wu2021hanet,han2021visual}. The general formula is denoted by
\begin{equation}
\begin{aligned}
\mathcal{L} =&\left[\Delta+S\left(V, T^{-}\right)-S(V,T)\right]_{+}\\
& +\left[\Delta+S\left(V^{-}, T\right)-S(V, T)\right]_{+} \label{triplet loss}
\end{aligned}
\end{equation}
where $\Delta$ represents the margin value for the pairwise ranking loss, $S$ denotes the similarity calculated by cosine similarity, $(V, T)$ is the positive pair, $V^{-}$and $T^{-}$are the hardest negatives. The operator $[x]_{+} = max(x, 0)$ is adopted twice to ensure that the matching samples should be close for both video-to-text and text-to-video retrieval.  

For choosing appropriate hard negative examples, most works~\cite{dong2019dual,mithun2018learning} adopt the hardest one that is closest to positive video-text pairs via the $argmax$ function. Besides, TCE~\cite{yang2020tree} demonstrates that choosing the hardest samples may lead to unstable and slow training in large batches, so they choose the top-5 samples to average the cost.
In addition, Mithun et al.~\cite{mithun2018learning} introduce a weighting function $L(r_{v/t})=1+1/(N-r+1)$ before each $[.]_{+}$ formula to make positive samples be top-ranking, where $r_{t}$ and $r_{v}$ denote the rank of matching text $t$ or video $v$ in all the compared samples.

\subsection{Contrastive Loss Functions}\label{Contrastive Loss}
Instead of inputting a triplet, contrastive loss takes positive and negative pairs as input. But its purpose and form are basically consistent with the triplet loss, guiding models to reduce the distance from positives and expand the distance from negatives.
For VTR, most studies~\cite{liu2021hit,sun2019learning,wang2022hybrid,ma2022x} are dedicated to multi-modal contrastive learning and introduce several loss functions.

\textbf{Symmetric cross entropy loss.}
Cross entropy loss functions are commonly used in the image classification task, while its learning process is very inconsistent for different categories. For some categories, it may quickly overfit to the wrong labels, while the under-fitting state may occur for other categories. Symmetric Cross Entropy~\cite{fang2021clip2video,luo2021CLIP4clip} is proposed to alleviate this problem. However, unlike the classification loss function, the formula in the VTR task is as follows:
\begin{equation}
\begin{gathered}
\mathcal{L}_{v 2 t}=-\frac{1}{B} \sum_{i}^{B} \log \frac{\exp \left({S}\left(v_{i}, t_{i}\right)\right)}{\sum_{j=1}^{B} \exp \left({S}\left(v_{i}, t_{j}\right)\right)} \\
\mathcal{L}_{t 2 v}=-\frac{1}{B} \sum_{i}^{B} \log \frac{\exp \left({S}\left(v_{i}, t_{i}\right)\right)}{\sum_{j=1}^{B} \exp \left({S}\left(v_{j}, t_{i}\right)\right)} 
\label{symmetric cross entropy}
\end{gathered}
\end{equation}
where $B$ is the batch size.
Normalized softmax loss (NSL)~\cite{qiu2019learning,zhai2018classification,bain2021frozen} owns the similar formula with an additional temperature hyper-parameter $\tau$, which is generally obtained by applying the features generated by FC to the Softmax activation function, followed by cross-entropy loss.

\textbf{Dual softmax loss} (DSL) is proposed in
CAMoE\cite{cheng2021improving} to make sure when the similarity sore from T2V is high, the score of V2T should be high as well, which is based on symmetric cross-entropy loss (Equ.~\ref{symmetric cross entropy}) and introduces a prior matrix on each $exp$ function. The prior matrices $\operatorname{Pr}^{v 2 t}$ and $\operatorname{Pr}^{t 2 v}$ are shown as followed respectively:
\begin{equation}
\begin{aligned}
&\operatorname{Pr}_{i, j}^{v 2 t}=\frac{\exp \left(l \cdot \operatorname{S}\left(v_{i}, t_{i}\right)\right)}{\sum_{j=1}^{B} \exp \left(l \cdot \operatorname{S}\left(v_{j}, t_{i}\right)\right)} \\
&\operatorname{Pr}_{i, j}^{t 2 v}=\frac{\exp \left(l \cdot \operatorname{S}\left(v_{i}, t_{i}\right)\right)}{\sum_{j=1}^{B} \exp \left(l \cdot \operatorname{S}\left(v_{i}, t_{j}\right)\right)}\label{DSL}
\end{aligned}
\end{equation}
where $l$ denotes a logit scaling parameter.
Since the proposal of the prior matrix, the single-side high score may not appear again so as to reduce computation and achieve higher performance. Especially in work~\cite{gao2021clip2tv}, the retrieval results improve 4.6\% performance on MSR-VTT 1k split R@1 metric due to the introduction of DSL.

\textbf{InfoNCE Loss.} In essence, Noise-Contrastive Estimation (NCE) Loss transforms the multiple classification problem solved by Softmax into a binary classification problem, which infers the real distribution via comparing the noisy samples with the noisy samples. Since the presence of multiple positive and negative samples in the VTR task resembles multiple categories, InfoNCE is proposed ~\cite{van2018representation} as a self-supervised contrastive loss~\cite{liu2021hit,gao2021clip2tv,ge2022miles,ma2022x}, which is formulated by:
\begin{equation}
\mathcal L =-log\frac{exp\left(S^{vt+}/\tau \right)}{\sum_{i=1}^{1+K_{t}}exp\left(S_{i}^{vt}/\tau \right)}-log\frac{exp\left(S^{tv+}/\tau \right)}{\sum_{i=1}^{1+K_{v}}exp\left(S_{i}^{tv}/\tau \right) }\label{InfoNCE}
\end{equation}
where $S^{vt+}$ is the positive similarity generated by cosine similarity. After calculating NCE loss, MILEs~\cite{ge2022miles} utilizes $l_{2}$ regressive objective loss to revise forecast features on the MVM module.

\subsection{Other Loss Functions}\label{Other loss}
\textbf{Distillation Loss.}
TeachText~\cite{croitoru2021teachtext} defines a novel similarity matrix distillation loss to convey the similarity matrix consistent with that from the student. Hence, the distillation loss is defined as:
\begin{equation}
\mathcal{L}_{d}=\frac{1}{B} \sum_{i=1}^{B} \sum_{j=1}^{B}\left[l\left(\Phi(i, j), S_{s}(i, j)\right)\right]
\end{equation}
where $\Phi=\Phi\left(S_{1}, \ldots, S_{N}\right)=$ $\frac{1}{N} \sum_{k=1}^{N} S_{k}$ is the aggregation of teacher similarity matrices and $S_{s}$ represents student similarity matrix. $l$ denotes the Huber loss, which is defined as:
\begin{equation}
l(x, y)= \begin{cases}\frac{1}{2}(x-y)^{2} & \text { if }|x-y| \leq 1, \\ |x-y|-\frac{1}{2} & \text { otherwise }\end{cases}
\end{equation}

\textbf{Cross-Modal Cycle Consistency Loss.}
COOT~\cite{ging2020coot} presents a cross-modal cycle consistency loss for better cross-modal alignment, which can be adopted in both supervised and unsupervised scenarios. To be specific, taking a sequence of video embeddings $\left\{\vartheta_i\right\}_{i=1}^n=$ $\left\{\vartheta_1, \ldots, \vartheta_n\right\}$ and sentence embeddings $\left\{\delta_i\right\}_{=1}^m=\left\{\delta_1, \ldots, \delta_m\right\}$ as input, finding soft nearest neighbors from both video-to-sentence and sentence-to-video direction. The loss aims to enforce the model to learn semantically cycle consistent representations via realizing the bidirectional soft location $i=\mu$.
\begin{equation}
\centering
\begin{aligned}
\bar{\vartheta}_{\delta_i} =\sum_{j=1}^n \alpha_j \vartheta_j , \quad
\alpha_j &=\frac{\exp \left(-\left\|\delta_i-\vartheta_j\right\|^2\right)}{\sum_{k=1}^n \exp \left(-\left\|\delta_i-\vartheta_k\right\|^2\right)}\\
\mu =\sum_{j=1}^m \beta_j j, \quad
\beta_j  & =\frac{\exp \left(-\left\|\bar{\vartheta}-\delta_j\right\|^2\right)}{\sum_{k=1}^m \exp \left(-\left\|\bar{\vartheta}-\delta_k\right\|^2\right)}\\
\mathcal{L}_{CMC}&  = \Vert i-\mu \Vert^{2}
\end{aligned}
\end{equation}
where $\alpha_j$ represents the similarity score of clip $\vartheta_j$ from sentence $\delta_i$.

\begin{table*}
\centering
\begin{adjustbox}{width=\textwidth}
\begin{threeparttable}
\caption{\large Commonly used datasets for video-text retrieval}\label{dataset}
\begin{tabular}{ c c c c c c c c c} \toprule
Dataset  &\#Videos  &\#Clips  & \#Captions & Duration &Avg.length &Source  &Caption source 
\\
\hline
HowTo100M~\cite{miech2019howto100m} &1.221M  &136M &136M &134,472h &6.5min &YouTube  &Manual\&Automatical 
\\ 
MSR-VTT~\cite{xu2016msr}  &7,180 &10k  &200k  &41.2h &20s &YouTube &Manual 
\\
ActivityNet Captions~\cite{krishna2017dense}    &20k  &100k  &100k &849h &180s &ActivityNet~\cite{caba2015activitynet} &Automatical   
\\
LSDMC~\cite{rohrbach2017movie}  &202 &118,081 &118,114  &158h &45s  &Movie &Manual 
\\ 
Vatex~\cite{wang2019vatex} &41.3k  &41.3k &826k &-&10s &Kinetics-600+YouTube &Manual 
\\
\bottomrule
\end{tabular}
\end{threeparttable}
\end{adjustbox}
\end{table*}

\section{Datasets and Results}\label{Datasets and Results}
In this section, we introduce some commonly-used datasets and report retrieval results on several standard metrics for comparison and analysis.

\subsection{Datasets}
\label{Datasets}
Some commonly-used datasets, which consist of numerous video clips and their corresponding annotated descriptions, are described in detail below. Their comparison is in Table~\ref{dataset}.

\textbf{MSR-VTT}~\cite{xu2016msr} includes 10k clips from 7180 videos composited by 20 different domains from the YouTube website, which is constructed to address the problem of translating videos to texts. 20 different sentences are annotated manually by 1327 workers and generate 20k clip-sentence pairs. Each clip lasts between 10 and 30 seconds, and the whole dataset takes over 40 hours. In addition, the audio information is collected into this dataset.
Generally, the test/train splits own three types: full split, 1k-A split, and 1k-B split. For the full split, there are 7k videos for training and 3k videos for testing. The 1k-A split utilizes 9k videos for training and 1k videos for testing and the 1k-B split adopts 6.5k videos for training and 1k videos for testing. The training videos are from the original train part, and the rest test part and 1k testing videos are selected from the test part of the full split. Besides, there is only one caption according to each video clip for 1k-A and 1k-B splits. At present, the 1k-A split is the most common way for comparing performance, see Table~\ref{MSR-VTT 1k-A results}.

\textbf{ActivityNet Captions}~\cite{krishna2017dense} consist of 20k video clips and 100k descriptions from ActivityNet. Each clip on average contains about 180 seconds for a total of 849 hours. Each caption contains an average of 13.48 words per sentence which is temporally localized and annotated via Amazon Mechanical Turk. At first, this dataset is collected for dealing with detecting and describing the overlap events that may concurrent in the same video simultaneously. Besides, this dataset references different categories, which means open for every domain. We follow the common evaluation setting~\cite{zhang2018cross,gabeur2020multi} to compare the results on the `val1' split.

\textbf{LSMDC} (Large Scale Movie Description Challenge)~\cite{rohrbach2017movie} contains 118,081 video clips and 118114 sentences with 158 hours of videos extracted from 202 distinct movies, and each video clip is around 45 seconds. The dataset is a combination of M-VAD and MPII-MD datasets. The former includes 46,589 clips and 55,904 sentences, and the latter dataset contains 68,337 clips and 68,375 sentences. In addition to the two datasets above, over 20 movie videos are collected consisting of 9,578 clips as the blind test, which is only used for evaluation. The remaining 118k clip-sentence pairs are split into a training set with 91,908 video clips, a validation set containing 6,542 clips, and a public test with 10,053 clips.

\textbf{Vatex}~\cite{wang2019vatex} consists of more than 41,250 videos with 825,000 Chinese and English subtitles. In the subtitles, there are more than 206,000 parallel English-Chinese translations.
Typically, 25,991 video clips are used for training, 1500 video clips for validation, and 1500 video clips for testing, where the validation and test sets are obtained by randomly splitting the official validation set of 3000 clips into two equal parts.

\textbf{HowTo100M}~\cite{miech2019howto100m} contains 136 million video clips sourced from 1.22 million narrated instructional videos on YouTube with 12 categories and referenced to 23.6k visual tasks (\eg, Food and Entertaining, Hobbies and Crafts, and Sports and Fitness). Each video clip lasts around 6.5 minutes, and the dataset takes up 134,472 hours. The 136 million annotations are generated via automatically transcribed narration by YouTube ASR or translated by YouTube API. At present, the HowTo100M dataset is commonly applied as a pre-trained dataset and then transferred to other fine-tuning datasets (\eg, MSR-VTT or LSMDC).

\begin{table*}
\centering
\begin{adjustbox}{width=\textwidth}
\begin{threeparttable}
\caption{Retrieval results on MSR-VTT 1k-A split. The second part introduces the methods based on the large-scale pre-trained model--CLIP, and the methods in the first part are based on other backbones.}\label{MSR-VTT 1k-A results}
\setlength{\tabcolsep}{1.6mm}{
\begin{tabular}{l| l|  c c c c c| c c c c c } \toprule
&\multirow{2}{1cm}{Method}      &\multicolumn{5}{ c| }{Text$\rightarrow$Video} &\multicolumn{5}{ c }{Video$\rightarrow$Text}\\

~ & ~&R@1$\uparrow$    &R@5$\uparrow$              &R@10$\uparrow$       & MdR$\downarrow$     &MnR$\downarrow$
 &R@1$\uparrow$    &R@5$\uparrow$              &R@10$\uparrow$       & MdR$\downarrow$     &MnR$\downarrow$ \\
\hline
\multirow{25}{1cm}{Others}
&Fusion~\cite{mithun2018learning}  &7.0 &20.9 &29.7 &38.0 &213.8  &12.5 &32.1 &42.4 &16.0 &134.0  \\
&JSFusion~\cite{yu2018joint} &10.2 &31.2 &43.2 &13.0 &- &- &- &- &- &-\\
&HT MIL-NCE~\cite{miech2019howto100m} &14.9 &40.2 &52.8 &9.0&- &- &- &- &- &-\\
&ActBERT~\cite{zhu2020actbert} &16.3 &42.8 &56.9 &10.0 &- &- &- &- &- &-\\
&HERO~\cite{li2020hero} &16.8 &43.4 &57.7 &-&- &- &- &- &- &-\\
&TCE~\cite{yang2020tree} &17.1 &39.9 &53.7 &9 &- &- &- &-&-&-\\
&Vidtranslate~\cite{korbar2020video} &17.4 &41.6 &53.6 &8.0&- &- &- &- &- &-\\
&ViSERN  &18.1 &48.4 &61.3 &6 &28.6 &24.3 &46.2 &59.5 &7 &34.6   \\ 
&CE~\cite{liu2019use} &20.9 &48.8 &62.4 &6.0 &28.2 &20.6 &50.3 &64.0 &5.3 &25.1\\
&UniVL~\cite{luo2020univl} &21.2 &49.6 &63.1 &6.0  &- &- &- &- &- &-\\
&ClipBERT~\cite{lei2021less} &22.0 &46.8 &59.9 &6.0  &- &- &- &- &- &-\\
&SSAN~\cite{guo2021ssan} &24.4 &49.3 &- &- &- &- &- &- &- &-\\
&MMT~\cite{gabeur2020multi}   &24.6 &54.0 &67.1 &4.0 &26.7 &24.4 &56.0 &67.8 &4.0 &23.6\\ 
&HCQ~\cite{wang2022hybrid} &25.9 &54.8 &69.0 &5.0 &- &26.3 &57.0 &70.1 &4.0 &-\\
&Ali et al.~\cite{ali2022video} &26.0 &56.7&-&3.0 &-&26.7 &56.5 &-&3.0&-\\
&MMT-pretrained~\cite{gabeur2020multi}  &26.6 &57.1 &69.6 &4.0 &24.0 &27.0 &57.5 &69.7 &3.7 &-\\
&SUPPORT-SET~\cite{patrick2020support} &27.4 &56.3 &67.7 &3.0 &- &26.6 &55.1 &67.5 &3.0 &-\\
&RIVRL~\cite{dong2022reading} &27.9 &59.3 &71.2 &4.0 &42.0 &- &- &- &- &-\\
&HiT \cite{liu2021hit} &28.8 &60.3 &72.3 &-&3 &27.7 &59.2 &72.0 &-&3\\  
&T2VLAD\cite{wang2021t2vlad}  &29.5 &59 &70.1 &4 &- &31.8 &60 &71.1 &3 &-\\ 
&TEACHTEXT\cite{croitoru2021teachtext} &29.6 &61.6 &74.2 &3 &- &- &- &- &- &-\\
&FROZEN~\cite{bain2021frozen} &31.0 &59.5 &70.5 &3.0 &- &- &- &- &- &-\\
&VSR-Net\cite{han2021visual}&37.2 &73.8 &85.0&- &2 &40.2 &76.1 &86.2 &- &2 \\ 
&BridgeFormer\cite{ge2022bridgeformer} &37.6 &64.8 &75.1 &3.0 &- &33.2 &58.0 &68.6 &4.0 &25.7\\
&MILES~\cite{ge2022miles} &37.7 &63.6 &73.8 &3.0 &- &- &- &- &- &-\\

\hline
\multirow{12}{1cm}{CLIP-based}
&CLIP~\cite{radford2021learning} &31.2 &53.7 &64.2 &4.0 &- &27.2 &51.7 &62.6 &5.0 &-\\
&MDMMT~\cite{dzabraev2021mdmmt} &38.9 &69.0 &79.7 &2.0 &16.5 &- &- &- &- &-\\

&CLIP4Clip~\cite{luo2021CLIP4clip}&44.5 &71.4 &81.6 &2.0 &15.3 &42.7 &70.9 &80.6 &2.0 &-\\
&CLIP2Video\cite{fang2021clip2video} &45.6 &72.6 &81.7 &2.0 &14.6 &43.5 &72.3 &82.1 &2.0 &10.2\\
&X-Pool~\cite{gorti2022x} &46.9 &72.8 &82.2 &2.0 &14.3 &- &- &- &- &-\\
&CAMoE
~\cite{cheng2021improving}   &47.3 &74.2 &84.5 &2.0 &11.9 &49.1 &74.3 &84.3 &2.0 &9.9\\
&CenterCLIP~\cite{zhao2022centerclip} &48.4 &73.8 &82.0 &2.0 &13.8 &47.7 &75.0 &83.3 &2.0 &10.2\\
&MDMMT-2~\cite{kunitsyn2022mdmmt} &48.5 &75.4 &83.9 &2.0&13.8 &-&- &- &- &- \\
&X-CLIP~\cite{ma2022x}  &49.3 &75.8 &84.8 &2.0 &12.2 &48.9 &76.8 &84.5 &2.0 &8.1\\
&CLIP2TV~\cite{gao2021clip2tv}  &52.9 &78.5 &86.5 &1.0 &12.8 &54.1 &77.4 &85.7 &1.0 &9.0\\
&DRL~\cite{wang2022disentangled} &53.3 &80.3 &87.6 &1.0 &- &56.2 &79.9 &87.4 &1.0 &-\\
 &HunYuan\_tvr~\cite{min2022hunyuan_tvr} &55.0 &80.4 &86.8 &1.0 &10.3 &55.5 &78.4 &85.8 &1.0 &7.7
\\
\bottomrule
\end{tabular}}
\end{threeparttable}
\end{adjustbox}
\end{table*}

\begin{table*}
\centering
\begin{minipage}{\textheight}
\small
\begin{threeparttable}
\caption{\large Retrieval results on three benchmarked datasets: ActivityNet Caption, LSMDS and Vatex.}\label{ results}
\setlength{\tabcolsep}{1.6mm}{
\begin{tabular}{ l  c c c c c c c c c c } \toprule
\multirow{2}{1cm}{Method}      &\multicolumn{5}{ c }{Text$\rightarrow$Video} &\multicolumn{5}{ c }{Video$\rightarrow$Text}\\

~  &R@1$\uparrow$    &R@5$\uparrow$              &R@10$\uparrow$       & MdR$\downarrow$     &MnR$\downarrow$
 &R@1$\uparrow$    &R@5$\uparrow$              &R@10$\uparrow$       & MdR$\downarrow$     &MnR$\downarrow$ \\
\hline
~ &\multicolumn{10}{ c }{Retrieval performance on ActivityNet Caption~\cite{krishna2017dense}}\\
\hline
MMT\cite{gabeur2020multi}&22.7 &54.2 &93.2 &5.0 &20.8 &22.9 &54.8 &93.1 &4.3 &21.2\\
HiT~\cite{liu2021hit} &27.7 &58.6 &94.7 &4.0& &&&&&\\
T2VLAD~\cite{wang2021t2vlad}&23.7 &55.5 &93.5 &4 &&24.1 &56.6 &94.1 &4&\\
COOT~\cite{ging2020coot} &60.8 &86.6 &98.6 &1&  &60.9 &87.4 &98.6 &1&\\
CLIP4Clip-meanPooling~\cite{luo2021CLIP4clip} &40.5 &72.4 &98.1 &2.0 &7.4  &42.5 &74.1 &85.8 &2.0 &6.6\\
CE~\cite{liu2019use} &17.7 &46.6 &- &6.0 &-&- &- &- &- &-\\
SUPPORT~\cite{patrick2020support} &28.7 &60.8 &- &2.0 &- &- &- &- &- &-\\
DRL~\cite{wang2022disentangled} &46.2 &77.3 &88.2 &2.0 &- &45.7 &76.5 &87.8 &2.0 &-\\
X-CLIP~\cite{ma2022x} &46.2 &75.5&& 
&6.8 &46.4 &75.9&& &6.4\\
HunYuan\_tvr~\cite{min2022hunyuan_tvr} &57.3 &84.8 &93.1 &1.0 &4.0 &57.7 &85.7 &93.9 &1.0 &3.4\\
\hline
~ &\multicolumn{10}{c}{Retrieval performance on LSMDC~\cite{rohrbach2017movie}}\\
\hline
NoiseE~\cite{amrani2021noise} &6.4 &19.8 &28.4 &39.0 &-&-&- &- &- &- \\
JSFusion~\cite{yu2018joint} &9.1 &21.2 &34.1 &36.0 &- &-&- &- &- &- \\
MEE~\cite{miech2018learning} &10.1 &25.6 &34.6 &27&- &-&- &- &- &-\\
CE~\cite{liu2019use} &11.2 &26.9 &34.8 &25.3 &- &- &- &- &- &-\\
MMT~\cite{gabeur2020multi} &12.9 &29.9 &40.1 &19.3 &75.0 &12.3 &28.6 &38.9 &20.0 &76.0\\
Frozen~\cite{bain2021frozen} &15.0 &30.8 &39.8 &20.0 &-&-&-&- &- &-  \\
TeachText-CE+~\cite{croitoru2021teachtext} &17.2 &36.5 &46.3 &13.7 &-&-&-&- &- &-  \\
MILES~\cite{ge2022miles}&17.8 &35.6 &44.1 &15.5&- &- &- &- &- &-\\
MDMMT~\cite{dzabraev2021mdmmt} &18.8 &38.5 &47.9 &12.3 &58.0&-&-&- &- &-  \\
CLIP4Clip-meanP~\cite{luo2021CLIP4clip}  &20.7 &38.9 &47.2 &13.0 &65.3&- &- &- &- &-\\
CLIP4Clip-seqTransf~\cite{luo2021CLIP4clip} &22.6 &41.0 &49.1 &11.0 &61.0 &20.8 &39.0 &48.6 &12 &54.2\\
X-Pool~\cite{gorti2022x} &25.2 &43.7 &53.5 &8.0 &53.2 &- &- &- &- &-\\
CAMoE~\cite{cheng2021improving} &25.9 &46.1 &53.7 &- &54.4 &- &- &- &- &-\\
X-CLIP~\cite{ma2022x} &26.1 &48.4&& 
&46.7 &26.9 &46.2&& &41.9\\
MDMMT-2~\cite{kunitsyn2022mdmmt} &26.9 &46.7 &55.9 &6.7 &48.0 &- &- &- &- &-\\
DRL~\cite{wang2022disentangled} &26.5 &47.6 &56.8 &7.0& - &27.0 &45.7 &55.4 &8.0 &-\\

HunYuan\_tvr~\cite{min2022hunyuan_tvr} &29.7 &46.4 &55.4 &7.0 &56.4 &30.1 &47.5 &55.7 &7.0 &48.9\\


\hline
~ &\multicolumn{10}{c}{Retrieval performance on Vatex~\cite{wang2019vatex}}\\
\hline
HANet~\cite{wu2021hanet} &36.4 &74.1 &84.1 &2& &49.1 &79.5 &86.2 &2&\\
VSE~\cite{faghri2017vse++} &28.0 &64.3 &76.9 &3.0 &- &- &- &- &- &-\\
SE++ &33.7 &70.1 &81.0 &2.0 &- &- &- &- &- &-\\
Dual Enc.~\cite{dong2019dual} &31.1 &67.5 &78.9 &3.0 &- &- &- &- &- &-\\
HGR~\cite{chen2020fine} &35.1 &73.5 &83.5 &2.0 &- &- &- &- &- &-\\
CLIP~\cite{radford2021learning}  &39.7 &72.3 &82.2 &2.0 &12.8 &52.7 &88.8 &94.9 &1.0 &3.8\\
SUPPORT-SET~\cite{patrick2020support} &44.9 &82.1 &89.7 &1.0 &- &58.4 &84.4 &91.0 &1.0 &-\\
CLIP4Clip-seqTransf~\cite{luo2021CLIP4clip} &55.9 &89.2 &95.0 &1.0 &3.9 &73.2 &97.1 &99.1 &1.0 &1.7\\
CLIP2Video~\cite{fang2021clip2video}&57.3 &90.0 &95.5 &1.0 &3.6 &76 &97.7 &99.9 &1.0 &1.5\\
DRL~\cite{wang2022disentangled}&65.7 &92.6 &96.7 &1.0 &-&80.1 &98.5 &99.5 &1.0&-\\
\bottomrule
\end{tabular}}
\begin{tablenotes}
\footnotesize
\item[].
\end{tablenotes}
\end{threeparttable}
\end{minipage}
\end{table*}

\subsection{Evaluaiton Metrics}\label{Evaluaiton Metrics}
The standard evaluation metrics for VTR include recalled at rank N (R@N, higher is better), median rank (MdR, lower is better), and mean rank (MnR, lower is better).

Recall refers to the ratio of the number of partially correct matches retrieved for relevant instances to the corresponding total number of given recalls. The vast majority of works employ R@1, R@5, and R@10 for assessment.
\begin{equation}
 Recall@K = \frac{correct~matches}{top-k~results}
\end{equation}

MdR calculates the median of ground-truth results in the ranking. MnR measures the average rank of all correct results.

\subsection{Performance Comparison 
}\label{Performance Comparison}

\textbf{Overview.} We present the performance results on MSR-VTT 1k-A split in Table~\ref{MSR-VTT 1k-A results} and other datasets (\ie, ActivityNet Caption, LSMDC, and Vatex) in Table~\ref{ results}. With the development of Transformer, most methods adopt pure Transformer network (\ie, ViT and BERT) to capture frame-wise or word-wise features~\cite{bain2021frozen,cheng2021improving}. As we can see in the last column of Table~\ref{MSR-VTT 1k-A results}, the methods based on CLIP achieve state-of-the-art performance in recent years~\cite{luo2021CLIP4clip,zhao2022centerclip,min2022hunyuan_tvr}. After that, several aggregation strategies (pooling, RNNs, CNNs, or attention mechanism~\cite{min2022hunyuan_tvr}) are utilized to fuse contextual information and generate fused feature representation. Feature embedding and matching are then performed to align video-text representation and generate the ranking list. Different loss functions are explored to optimize the ranking. In the past two years, the performance has improved with a margin of over 20\% on the MSR-VTT R@1 metric (See Table~\ref{MSR-VTT 1k-A results}). 

\textbf{Evaluation on multi-modal video representation.} For capturing considerable knowledge contained in videos, existing works utilize multiple experts~\cite{liu2019use,gabeur2020multi,wang2021t2vlad,liu2021hit} to pre-extract different modal information on raw videos (See the first column in Tab.~\ref{classify}), and then aggregate the extracted features by max pooling, clustering center, or gating mechanism to obtain video representations. These video features learn comprehensive information obtained after fully mining various modalities in videos. To be specific, MEE~\cite{miech2018learning} extracts features of motion, audio, face, and appearance as video representation. Fusion~\cite{mithun2018learning} chooses RGB, motion, and audio feature extractors to generate representation. The importance of each modality has been examined in work~\cite{gabeur2020multi} as shown in Table~\ref{each expert}. It can be discovered that when extracting only one modality feature, motion, appearance or scene performs better than others with lower than 50\% results. Besides, it can be confirmed that the audio features as supplementary information make the features more complete under the interaction of multiple modalities. However, it is observed that in the last column, the combination of too many modalities leads to performance degradation, indicating that the considered modal information may cause noise impact on VTR. Therefore, for the datasets with audio information such as MAR-VTT and LSMDC, most works pre-extract visual features as well as motion and audio features, while some other studies only extract appearance features. 

\begin{table}[ht]
    \centering
      \caption{Performance evaluation of each expert. Numerical results cited from work~\cite{gabeur2020multi} are based on the mean rank evaluation metric on the MSR-VTT dataset. `w' represents only extracting this modality as video features, `w/o' indicates capturing all modality features only without this one, and `Overlay' means results generated by adding corresponding modality features from top to bottom sequentially.} \label{each expert}
    \begin{threeparttable}
    \begin{tabular}{c|c|c|c}
   \toprule
        Expert & w MnR & w/o MnR & Overlay MnR\\
        \hline
         motion &35.9 &29.5 & 35.9\\ 
         audio &127.0 &30.0 &29.0\\
         appearance &40.7 &26.8 &27.6\\
         scene &50.0 &26.4 &26.5\\
         speech &330.2 &26.7 &26.1\\
         ocr&393.4 &27.0 &25.9\\
         face &299.8 &25.9 &26.7\\
         \hline
         ALL &26.7 &26.7 &26.7\\
    \botrule
    \end{tabular}
    \end{threeparttable}
\end{table}

\textbf{Evaluation on video spatio-temporal characteristics.} Nowadays, most methods of extracting video features are to extract spatial features first and then capture temporal information. As shown in Table~\ref{classify}, previous works employ pre-trained CNNs to capture static frame features as spatial features, and RNNs based methods for temporal encoding, similar to textual encoders. However, Dual encoding~\cite{dong2019dual} achieves only 7.7\% on the MSR-VTT full split R@1 metric and demonstrates the extracted video information is insufficient, leading to low accuracy.
Next, many approaches leverage CNNs to extract changeable information (\ie, motion and audio) along the temporal dimension, and the performance improves by over 10\% than the former way. Furthermore, with the flourish of Transformer on both video and text (especially ViT and BERT), the approaches based on the combination of CNNs and Transformer become a favorable fashion. MDMMT-2~\cite{kunitsyn2022mdmmt} has found that introducing three experts on CLIP improves the results by 0.9\% on the MSR-VTT full split R@1 metric. In recent years, more and more methods begin to adopt CLIP as a backbone and gradually break through state-of-the-art performance, which has improved dramatically by over 40\%, even reaching 55\% on the MSR-VTT 1k-A split R@1 metric. 

\textbf{Evaluation on multi-grained matching.}
In recent years, multi-grained matching has been sought after by most researchers. Table~\ref{grained evaluation} describes the retrieval performance obtained by several methods based on different fine-grained matching. Overall, the retrieval performance with only fine-grained matching is better than that with only global matching. Meanwhile, the combination of multiple fine-grained matching achieves the best performance. In particular, HANet~\cite{wu2021hanet} studies various fine-grained combinations on the MSR-VTT full split, and we can see the individual level matching achieves 8.6\%, and the combination of each level matching increases the accuracy with about 0.7\%. T2VLAD~\cite{wang2021t2vlad} demonstrates the superiority of local matching, whose performance increases with 2.1\% than global matching alone. Similarly, the difference between individual and global matching reaches 2.5\% in DRL~\cite{wang2022disentangled}. Additionally, VSR-Net~\cite{han2021visual} performs fine-grained matching and focuses on the alignment between proposal objects (which is generated from some local portions in frames and fused by an attention-aware feature aggregation layer) and captions. The combination increase 6.9\% in VSR-Net~\cite{han2021visual} and 7.3\% in T2VLAD, and 5.4\% in HunYuan\_tvr~\cite{min2022hunyuan_tvr}. In addition, DRL proposes a weighted token-wise interaction mechanism that emphasizes the different importance of each frame and word, which improves the performance by 1.1\%. VSR-Net introduces the combination of global embedding learning and fine-grained relation embedding learning, which are categorized as global and local matching in this survey.
\begin{table*}

    \centering
      \caption{Performance evaluation of multi-grained matching on the MSR-VTT 1k-A split. `\checkmark' represents that this hierarchical matching is adopted. `$\star$' means the performance of HANet is evaluated on another dataset split, \ie, full split.}\label{grained evaluation}
    \resizebox{\linewidth}{!}{ 
    \scriptsize
    \begin{threeparttable}
    \setlength\tabcolsep{3pt}
    \begin{tabular}{c|c|ccc|ccccc}
    
   \toprule
        \multirow{2}{0.6cm}{Method}&\multirow{2}{0.7cm}{Backbone} & \multicolumn{3}{c}{Multi-grained} &\multicolumn{5}{ c }{Text$\rightarrow$Video}\\
       ~ &~ &Global&Local&Individual&R@1$\uparrow$    &R@5$\uparrow$              &R@10$\uparrow$       & MdR$\downarrow$     &MnR$\downarrow$\\
        \hline
        \multirow{4}{*}{HANet~\cite{wu2021hanet}$\star$} &\multirow{4}{*}{ResNet152} &\checkmark&~&~ & 8.4 &24.9 &35.6 &-&24\\
         ~&&&\checkmark&~&8.4 &24.6 &34.9&-& 25\\
         ~&&&&\checkmark&8.6 &25.7 &36.4&- &23\\
         ~&&\checkmark&\checkmark&\checkmark&9.3 &27.0 &38.1&- &20\\
         \hline
         \multirow{3}{*}{T2VLAD~\cite{wang2021t2vlad}} &\multirow{3}{*}{Experts} &\checkmark&~&~ & 22.2&49.9&64.6&6&-\\
         ~&&&\checkmark&~&24.3&51.5&63.4&5&-\\
         ~&&\checkmark&\checkmark&&29.5&59.0&70.1&4&-\\
         \hline
          \multirow{3}{*}{VSR-Net~\cite{han2021visual}}&\multirow{3}{*}{InceptionResNetV2,I3D} &\checkmark&~&~ & 30.3&66.6&78.0&3&-\\
         ~&&&\checkmark&~&29.9&63.4&76.0&3&-\\
         ~&&\checkmark&\checkmark&&37.2&73.8&85.0&2&-\\
         \hline
         \multirow{3}{*}{DRL~\cite{wang2022disentangled}} &\multirow{3}{*}{CLIP(ViT-B/16)} &\checkmark&~&~ &46.6&73.3&82.8&-&13.4\\
        ~&&&&\checkmark&49.1&75.7&85.1&-&12.7\\
         ~&&&&\checkmark&50.2&76.5&84.7&-&12.4\\
         \hline
         \multirow{3}{*}{HunYuan\_tvr~\cite{min2022hunyuan_tvr}}&\multirow{3}{*}{CLIP(ViT-B/16)} &\checkmark&~&~ &44.6&72.0&82.3&2&13.6\\
        ~&&&&\checkmark &48.3&74.9&83.8&2&12.4\\
         ~&&\checkmark&\checkmark&\checkmark &49.0&76.5&84.3&2&11.8\\
    \botrule
    \end{tabular}
    \end{threeparttable}
    }
\end{table*}
 
\textbf{Evaluation on fine-tuning.}
Objective loss functions as an effective way of fine-tuning, refer to Section~\ref{Training Objectives}, can generally improve the retrieval performance compared with no fine-tuning methods. In Table~\ref{finetune evaluation}, we compare recent approaches based on the Transformer architecture. For instance, DSL improves T2V retrieval performance by 5\% on MSR-VTT dataset R@1 in work~\cite{min2022hunyuan_tvr}, and similar results are reported in 
works~\cite{cheng2021improving,gao2021clip2tv}.
In addition, choosing a more appropriate backbone or aggregation strategy may also improve the performance. The performance in work~\cite{gorti2022x} increased from 46.1\% to 49.3\% on MSR-VTT, from 23.3\% to 26.1\% on LSMDC, and improved by 2.8\% in work~\cite{wang2022disentangled} on MSR-VTT. Of course, there are cases where replacing the backbone fails to provide a significant performance boost. For example, the accuracy in Hunyuan\_tvr increases by only 0.2\% on MSR-VTT, or even decreased by 2.6\% on LSMDC, because it is a small-scale dataset, and large-scale models trained on it are prone to overfitting. Therefore, selecting an appropriate backbone is an important part. Due to the introduction of the multi-head attention mechanism, the feature fusion mechanism is further developed. The results have increased by over 1\% both on MSR-VTT and LSMDC in work~\cite{luo2021CLIP4clip}. Besides, as we can see, although X-CLIP and DRL utilize the same backbone, attention aggregation, and loss function, X-CLIP chooses 
the weighted sum of cross-level similarity matrices to obtain the results, and DRL proposes WTI to consider both token-wise and channel-wise similarities. Because of more finer-grained matching and denoising process, the performance of DRL is slightly higher than X-CLIP.

\begin{table*}
    \centering
      \caption{Performance evaluation of methods in which Transformer based models are fine-tuned. All of the loss functions can be found in Section~\ref{Training Objectives}. The accuracy numbers in parentheses represent the performance generated without employing the aforementioned loss function. All performance numbers are obtained by the T2V retrieval task on the R@1 metric. 
      }\label{finetune evaluation}
    \resizebox{\linewidth}{!}{ 
    \begin{threeparttable}
    \begin{tabular}{c|c|c|c|c|c}
   \toprule
        Method & Backbone & \makecell{Embedding\\ Aggregation} & \makecell{Loss\\ function} &MSR-VTT 1k-A & LSMDC\\
        \hline
         CLIP4Clip~\cite{luo2021CLIP4clip} &CLIP (ViT-B/32) & \makecell{meanP\\SeqLSTM\\SeqTransf\\tightTransf} &CE loss&\makecell{43.1\\42.5\\44.5\\40.5}&\makecell{20.7\\21.6\\22.6\\18.9}
         \\
         \hline
         X-CLIP~\cite{ma2022x} &\makecell{CLIP(ViT-B/32)\\CLIP(ViT-B/16)}&attention &InfoNCE &\makecell{46.1\\49.3} &\makecell{23.3\\26.1} \\
         \hline
         DRL~\cite{wang2022disentangled} &\makecell{CLIP(ViT-B/32)\\CLIP(ViT-B/16)}&attention &InfoNCE &\makecell{47.4\\50.2}&\makecell{24.9\\26.5}\\
         \hline
         CenterCLIP~\cite{zhao2022centerclip} & CLIP4Clip(ViT-B/16)-meanP &k-medoids++& NSL &48.4&24.2\\
         \hline
         X-Pool~\cite{gorti2022x} &CLIP(ViT-B/32) &\makecell{top-k pool\\mean pool}&CE loss & \makecell{42.1\\44.6}&-\\
         \hline
          CAMOE~\cite{cheng2021improving} & ViT-B/32+BERT &attention& DSL &47.3(44.6)&25.9(22.9)\\ \hline
         CLIP2TV~\cite{gao2021clip2tv}&CLIP(ViT-B/16) &MMT& DSL&52.9(48.3)&-\\
         \hline
         Hunyuan\_tvr~\cite{min2022hunyuan_tvr} &\makecell{CLIP(ViT-L/14)\\CLIP(ViT-B/16)} &attention&DSL&\makecell{-(49.5)\\54.9(49.7)}&\makecell{29.7(27.1)\\-(24.5)}\\
    \botrule
    \end{tabular}
    \end{threeparttable}}
\end{table*}

\section{Challenges and Directions}\label{Challenges and Directions}
\subsection{Major Challenges}\label{key challenges}
The video-text retrieval task has achieved rapid progress in recent years. In particular, the retrieval result of the R@1 metric on the MSR-VTT dataset has even improved from 24.1 to 55.0 in two years (See Tab.~\ref{MSR-VTT 1k-A results} for details). However, some inherent challenges are still remaining:
\begin{itemize}
    \item [(1)]\textbf{Video feature extraction} needs to capture rich and comprehensive features contained in videos, such as spatio-temporal, and multiple modalities information. Previous studies~\cite{miech2018learning,gabeur2020multi,wang2021t2vlad} consider the significance of modalities and choose several feature extractors to extract different modality features in videos as `expert'. 
    Significantly, the choice of modal category is also an issue that may affect the performance (See the performance evaluation on multi-modal video representation in Sec.~\ref{Performance Comparison} for details). 
    Several works~\cite{miech2018learning,yu2018joint} capture each sample frame's feature and then integrate them as a video representation, whereas they ignore the temporal correlations between two successive frames. In recent years, most works~\cite{ma2022x,gao2021clip2tv} are devoted to capturing spatial and temporal features jointly. The prosperity of Transformer provides spatio-temporal joint learning architectures~\cite{bain2021frozen,bertasius2021space,arnab2021vivit} to extract features simultaneously. Besides, several studies~\cite{kunitsyn2022mdmmt} combine a few experts with spatio-temporal feature extractors and gain some performance improvements. Besides, Kunitsyn et al.~\cite{kunitsyn2022mdmmt} prove that the performance of only applying audio and motion experts to extract features underperforms the combination of CLIP and these two experts, by about 8\% on the MSR-VTT 1k-A split. In other words, the performance comparison in this paper confirms that the feature extractor based on the CLIP model plays a dominant role, while the experts are generally used to supplement the existing model. Nevertheless, extracting complete and robust video features still faces challenges, which need to be further studied in the future.
    \item [(2)]\textbf{Cross-modal gap} has been a common and important problem for many multi-modal applications. It can be seen that the cross-modal gap between video and text retrieval becomes particularly large due to the information asymmetry, \ie, the sentences are generally very simple and concise, while the videos contain complicated and expensive information. Gabeur et al.~\cite{gabeur2020multi} propose the gated embedding for mapping text into the same dimension as each video modality to bridge the gap. In addition, the combination of global and local alignment methods fully considers coarse-grained and fine-grained feature interactions~\cite{wu2021hanet,min2022hunyuan_tvr}. According to Tab.~\ref{grained evaluation}, it is clear that combining multiple fine-grained matching strategies is superior to any single fine-grained one. Besides, as we can see in Fig.~\ref{1}, few works have studied the method of combining local and individual matching, while it still remains to be explored in the future. Nevertheless, it remains challenging about how to effectively remove the feature mis-matching caused by noisy information. Furthermore, most methods are trained with complete video-text pairs without considering the absence of either of the two modalities. Hence, weakly supervised methods still need to be explored for VTR.

    \item [(3)]\textbf{Efficiency.} 
    The successes of advanced deep learning approaches rely mostly on expensive training time and memory costs. However, it is impractical for many real-world applications. One potential solution is leveraging the pre-training model like CLIP to extract multi-modal representation~\cite{zhao2022centerclip,luo2021CLIP4clip}, with no need for training from scratch anymore. However, the drawback is that we cannot fine-tune the pre-trained CLIP for the target data due to its huge number of network parameters. Thereby, this may limit further performance improvements. For reducing computation, some studies remove fine-grained features matching~\cite{gabeur2020multi,guo2021ssan}. However, only focusing on global features may lead to performance degradation compared to combining with local features. The other way is to encode and compress simultaneously based on high-dimensional embeddings~\cite{wang2022hybrid}, while this method has not been widely studied in the video retrieval field and has not achieved high performance. Besides, considering memory cost, the high-dimensional deep feature is expensive and slow for retrieval tasks. Therefore, it is necessary to develop compact and low-dimensional feature representation extractors without sacrificing accuracy performance. One solution is to design new hashing strategies~\cite{gu2016supervised,wu2018unsupervised,zhuo2022clip4hashing} which are widespread in image and video retrieval for storage and searching while having not been  researched much for the VTR task.
\end{itemize}
    
\color{black}

\subsection{Future Directions}
\label{Future Directions and Outlook}
Video-text retrieval is a complex and challenging task, which not only has the difficulty of video feature extraction but also the cross-modal problem between videos and sentences. Although deep learning has made great achievements in feature extraction and matching in the past few years, there are still some potential directions to be explored further for VTR.



\textbf{(1) Accurate and compact video representation.} 
One of the ongoing challenges remains about how to adequately capture both multi-modal and spatio-temporal video representation, as mentioned in Section~\ref{key challenges}. Despite the proposal of Transformer-based approaches has achieved better performance than previous network architectures, however, it still lacks robustness and generalization in the wild. 
Hence, more discriminative and comprehensive features should be fully extracted to improve the accuracy of the retrieval task.  
Furthermore, we should avoid blindly applying a mixture of models to capture complete representations, as~Gabeur et al.~\cite{gabeur2020multi} demonstrate that considering too much modality information can also lead to performance degradation. Several researchers~\cite{mithun2018learning,gabeur2020multi} choose the appearance, audio, and motion modalities information that can achieve the best performance. In one word, we should further study an appropriate learning scheme to generate accurate and compact video features. 



\textbf{(2) Reduce redundancy and improve efficiency.} 
Recall that Transformer based approaches divide video frames into multiple patches owning the same size. Due to the redundancy of continuous and similar frames in one video, many homogeneous tokens are likely to be generated leading to increasing computational costs. Recent work~\cite{zhao2022centerclip} exploits clustering centers to reduce redundant tokens via suppressing non-essential content and reconstructing novel token sequences, utilizes condition pooling~\cite{gorti2022x} to aggregate visual tokens according to related text information, and even masks the unimportant tokens~\cite{rao2021dynamicvit}.
In addition, to improve retrieval efficiency, both quantization and knowledge distillation can be introduced to decrease computation~\cite{croitoru2021teachtext,wang2022hybrid}. Very recently, CLIP4hashing~\cite{zhuo2022clip4hashing} plants unsupervised deep hashing categories into the VTR task. However, its performance is much lower than state-of-the-art performance. Therefore, reducing the interaction of redundant information and achieving efficient retrieval still deserves further investigation.

 
\textbf{(3) Learn to retrieve with weak supervision.}
It is obvious to find that pre-training on large-scale datasets (\eg, HowTo100M) and fine-tuning on specific datasets has become a major pipeline for VTR. 
Recent work in~\cite{kunitsyn2022mdmmt} introduces HowTo100M and crowd-labeled datasets to train in different stages and combine them, which wins stronger generalization ability and can be transferred to multiple domains for retrieval without fine-tuning. However, the combination of numerous datasets to train is rarely researched. In addition, when training datasets are not combined properly (\eg, removing the overlap of the datasets~\cite{dzabraev2021mdmmt}), the performance may be degraded largely. To this end, we need to consider how to improve retrieval ability under weak supervision.
 

\section{Conclusion}
\label{conclusion}
In this survey, we have provided a comprehensive overview of deep learning methods for the video-text retrieval task. We have reviewed recent major advances on video-textual feature extraction, feature matching and objective functions, presented their performance on publicly benchmarked datasets, and analyzed their merits and limitations. Although the VTR task has achieved remarkable progress over recent years, it is worth mentioning that there are still some potential research space. Finally, we have summarized the difficulties and challenges faced by this task and discussed several future research directions,
with the expectation to provide new insights for researching video-text retrieval in the future.

\section*{Statements and Declarations}
The results/data/figures in this manuscript have not been published elsewhere, nor are they under consideration (from you or one of your Contributing Authors) by another publisher.
We have read the Springer journal policies on author responsibilities and submit this manuscript in accordance with those policies.
All of the material is owned by the authors and/or no permissions are required.

The authors have no competing interests to declare that are relevant to the content of this article.
The authors have no competing interests as defined by Springer, or other interests that might be perceived to influence the results and/or discussion reported in this paper.

The datasets generated during and/or analysed during the current study are available from the corresponding author on reasonable request.

\bibliography{example}

\end{document}